\documentclass[lettersize,journal]{IEEEtran}
\usepackage{amsmath,amsfonts}
\usepackage{algorithmic}
\usepackage{algorithm}
\usepackage{array}
\usepackage[caption=false,font=normalsize,labelfont=sf,textfont=sf]{subfig}
\usepackage{textcomp}
\usepackage{stfloats}
\usepackage{url}
\usepackage{verbatim}
\usepackage{cite}
\usepackage{multirow} %add package for merge table rows
\usepackage[normalem]{ulem}
\usepackage{float}
\usepackage{subcaption}
\usepackage{graphicx}
\usepackage[table]{xcolor}
\useunder{\uline}{\ul}{}
\usepackage{algorithm}
\usepackage{algorithmic}
\usepackage{amsmath}
\usepackage{hyperref}
\hypersetup{
    colorlinks=true,
    linkcolor=blue,
    filecolor=blue,      
    urlcolor=blue,
    citecolor=blue,
}
\usepackage{bm}
\usepackage{booktabs}
\usepackage[table]{xcolor}

\begin{document}

\title{EvoEdit: Lifelong Free-Text Knowledge Editing through Latent Perturbation Augmentation and Knowledge-driven Parameter Fusion}

\author{Pengfei Cao, Zeao Ji, Daojian Zeng, Jun Zhao, Kang Liu 
        % <-this % stops a space
\thanks{Pengfei Cao, Jun Zhao and Kang Liu are with the Key Laboratory of Cognition and Decision Intelligence for Complex Systems, Institute of Automation, Chinese Academy of Sciences, and the School of Artificial Intelligence, University of Chinese Academy of Sciences, Beijing, China. E-mails: \{pengfei.cao, jzhao, kliu\}@nlpr.ia.ac.cn.}% <-this % stops a space
\thanks{Zeao Ji and Daojian Zeng are with the Institute of AI and International Communication, Hunan Normal University, Changsha, China. E-mails: \{202420081572, zengdj\}@hunnu.edu.cn.}
\thanks{Corresponding authors: Daojian Zeng and Kang Liu. The first two authors contributed equally.}
}

% The paper headers
% \markboth{Journal of \LaTeX\ Class Files,~Vol.~14, No.~8, August~2021}%
% {Shell \MakeLowercase{\textit{et al.}}: A Sample Article Using IEEEtran.cls for IEEE Journals}

\maketitle

\begin{abstract}
Adjusting the outdated knowledge of large language models (LLMs) after deployment remains a major challenge. This difficulty has spurred the development of knowledge editing, which seeks to accurately and efficiently modify a model’s internal (parametric) knowledge without retraining it from scratch. However, existing methods suffer from two limitations. First, they depend on structured triplets that are misaligned with the free-text nature of LLM pretraining and fail to capture the nuanced relationships among facts. Second, they typically support one-time knowledge updates, with relatively limited research on the problem of sequential or lifelong editing. To address these gaps, we propose a new task, Lifelong Free-text Knowledge Editing (LF-Edit), which enables models to incorporate updates expressed in natural language and supports continual editing over time. Despite its promise, LF-Edit faces the dual challenge of integrating new knowledge while mitigating the forgetting of prior information. To foster research on this new task, we construct a large-scale benchmark, Multi-Rank Lifelong Free-text Editing Benchmark (MRLF-Bench), containing 16,835 free-text edit requests. We further design a cognitively inspired multi-rank evaluation framework encompassing four levels: memorization, understanding, constrained comprehension, and reasoning. To tackle the challenges inherent in LF-Edit, we introduce a novel approach named \textsc{EvoEdit} that enhances knowledge injection through Latent Perturbation Augmentation and preserves prior information via Knowledge-driven Parameter Fusion. Experimental results demonstrate that \textsc{EvoEdit} substantially outperforms existing knowledge editing methods on the proposed LF-Edit task.
\end{abstract}

\begin{IEEEkeywords}
Lifelong Free-text Knowledge Editing, Perturbation Augmentation, Parameter Fusion, Large Language Model.
\end{IEEEkeywords}

\section{Introduction}
\IEEEPARstart{L}{arge} Language Models (LLMs), trained on massive-scale pretraining corpora, have acquired extensive parametric knowledge and demonstrated remarkable capabilities in understanding and generating natural language. However, LLMs primarily rely on knowledge acquired during pretraining to generate responses, which suffers from critical limitations. Firstly, real-world knowledge evolves continuously, but the once-for-all training paradigm prevents models from updating their internal information. Once deployed, model parameters are fixed, updating the outdated knowledge is not an easy task \cite{zhang2024comprehensivestudyknowledgeediting}. Secondly, during pretraining, noisy and biased data are inevitably encoded into the model's parameters, causing LLMs to acquire erroneous facts and misleading information that cannot be easily corrected \cite{cheng2024editmultimodallargelanguage,survey_for_knowledge_editing}. To this end, several strategies have been proposed to update the knowledge encoded in LLMs, such as continual pretraining \cite{kecontinual}, supervised fine-tuning (SFT) \cite{survet_of_finetuning}, and retrieval-augmented generation (RAG) \cite{survey_of_RAG}. However, these approaches are often limited by their prohibitive cost, extensive time requirements, and high resource consumption \cite{grace}. Thus, knowledge editing has emerged as a promising solution that directly modifies parameters related to erroneous or outdated knowledge, enabling LLMs to efficiently update their internal knowledge.

\begin{figure}[t]
    \centering
    \includegraphics[width=0.97\columnwidth]{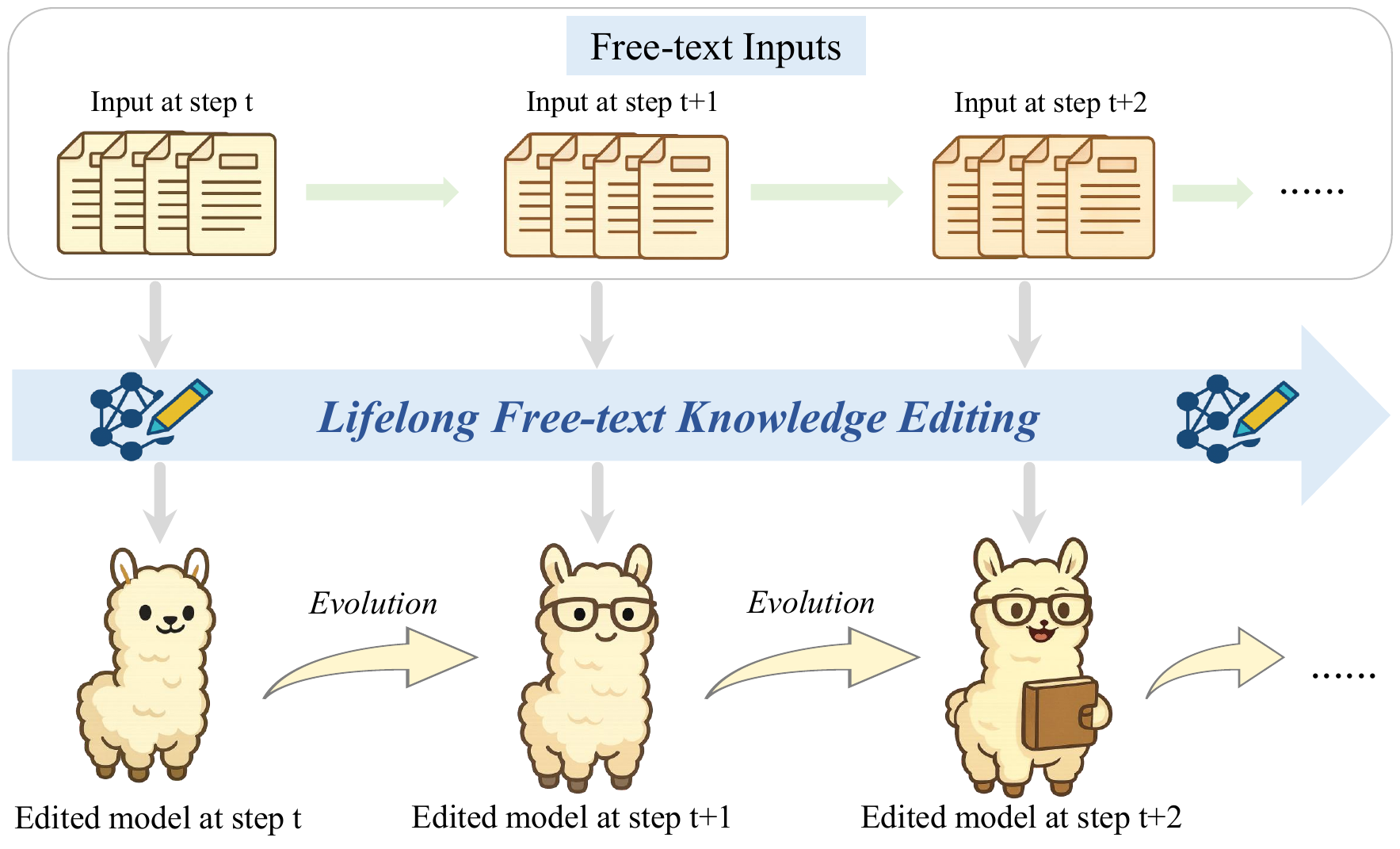}
    \caption{Overview of the lifelong free-text knowledge editing (LF-Edit) task. The target of the task is to enable LLMs to continuously acquire knowledge from free text and efficiently update the internal knowledge of the model.}
    \label{fig_1}
\end{figure}

Existing works mainly focus on structured inputs, typically represented as \textit{(subject, relation, object)} triples. In general, these methods can be classified into three categories. The first, Locate-Then-Edit based Methods, employ localization techniques to identify model components responsible for specific knowledge and then directly modify the corresponding parameters \cite{rome,r-ROME,memit,alphaedit,fine,hu2024wilke,cao2024one}. The second, Meta-Learning based Methods, leverage training data to train a hypernetwork that predicts parameter adjustments for encoding new knowledge \cite{ke,mend,malman,wang2024mulfe}. The final category, Memory based Methods, leaves the original model unchanged, instead storing new knowledge in an external memory module and retrieving it during inference \cite{serac, ike, grace, mello, wise}. Despite these efforts for knowledge editing, these methods exhibit two critical limitations:

\textbf{1) Confined to triplet-form inputs:} Most existing methods for knowledge editing rely on structured triplets. However, this structured representation often diverges significantly from the natural language inputs used during pretraining, which typically encompass multiple facts and their relations. Structured triplets are limited to encoding individual facts and fail to capture the complex relationships between them. This limitation results in incomplete or inaccurate knowledge injection, which in turn may degrade model performance. Furthermore, to adapt to existing knowledge editing methods, we need to extract structured facts from free text. However, the extraction tools are prone to various errors \cite{survey_for_extraction}, such as redundant extractions that generate multiple triplets for the same subject. This introduces noises into the editing process, ultimately undermining the effectiveness of triplet-based knowledge editing methods. \textbf{2) Restricted to one-time editing:} Existing knowledge editing methods are mainly designed for single-step updates, lacking research on sequential or lifelong editing. However, in real-world scenarios where knowledge continuously evolves, these single-step editing methods cannot integrate successive knowledge updates into the model. As a result, outdated information still persists in models, leading to a lack of correct perception of the external world, which ultimately undermines the reliability of the model output \cite{rome,memit,jin2024cutting}.

The limitations discussed above motivate us to propose a novel task: \textbf{L}ifelong \textbf{F}ree-text Knowledge \textbf{Edit}ing (\textbf{LF-Edit}), which enables models to incorporate updates in a more natural and continuous manner. As illustrated in Figure \ref{fig_1}, LF-Edit accepts free-text inputs, rather than structured triplets. By leveraging these inputs, LF-Edit aims to support lifelong knowledge editing in LLMs, facilitating the updating of outdated or incorrect information and thereby maintaining models' accuracy and timeliness. Despite its potential benefits, LF-Edit faces two fundamental challenges: \textbf{1) Injecting Knowledge with Free-text Form:} Using free text for knowledge injection is inherently challenging, as such text often contains fragmented knowledge and complex connections. It is challenging to inject all relevant knowledge into the model at once. The scattered and unstructured nature of free text complicates the comprehensive identification and precise integration of knowledge. Furthermore, relationships such as causal, temporal, and spatial dependencies are frequently implicit and dispersed across multiple sentences, making them difficult to be acquired fully during knowledge editing. \textbf{2) Catastrophic Forgetting:} In a lifelong setting, knowledge must be updated continuously, with each update introducing subtle shifts to the model’s internal representations. As the editing process continues, these shifts may interfere with previously acquired knowledge and gradually cause the model to deviate from its original state \cite{survey_for_knowledge_editing,hu2024wilke,hu2025knowledge}. This cumulative effect can ultimately degrade both the reliability and the overall performance of the model.

To promote the research of this new task, this paper develops a large-scale benchmark named \textbf{M}ulti-\textbf{R}ank \textbf{L}ifelong \textbf{F}ree-text Editing \textbf{Bench}mark (\textbf{MRLF-Bench}). Constructed from Wikidata, MRLF-Bench comprises time-sensitive entries whose factual attributes evolve across different temporal stages. The benchmark consists of 16,835 free-text instances specifically designed to capture real-world knowledge dynamics. Drawing inspiration from cognitive developmental psychology \cite{Piaget}, we evaluate model performance from multi-level analytical dimensions, including knowledge memorization, understanding, constrained comprehension, and reasoning. This multi-rank evaluation framework enables a fine-grained diagnosis of editing capabilities, ranging from surface-level recall to advanced reasoning.

Furthermore, we propose an effective method named \textbf{\textsc{EvoEdit}} to address the aforementioned challenges in this new task. \textsc{EvoEdit} comprises two core components, including \textit{Latent Perturbation Augmentation} and \textit{Knowledge-driven Parameter Fusion.} Specifically, to facilitate the acquisition of new knowledge from free-text inputs, the latent perturbation augmentation introduces controlled noise into the latent representations of LLMs, serving as a form of implicit data augmentation during the editing process. To mitigate catastrophic forgetting in lifelong editing, we further design a knowledge-driven parameter fusion strategy. This strategy first estimates parameter importance by measuring their correlation with knowledge edits and then selectively integrates the most relevant parameters. Through this design, \textsc{EvoEdit} effectively preserves prior knowledge while incorporating new information. Experimental results demonstrate that our approach surpasses existing knowledge editing methods and achieves superior performance on the novel LF-Edit task. 

Overall, the main contributions of this work can be summarized as follows:

\begin{itemize}
    \item We formally define a challenging task named lifelong free-text knowledge editing (LF-Edit), which enables LLMs to continuously acquire new knowledge from free-text inputs while preserving previously learned knowledge without forgetting.
    
    \item We introduce MRLF-Bench, a multi-rank benchmark for lifelong editing on free-text. It comprises 16,835 time-related edits, each accompanied by four-tier evaluation questions designed to assess knowledge memorization, understanding, constrained comprehension, and reasoning. This benchmark establishes a solid foundation for future research on lifelong free-text editing.
    
    \item We propose the \textsc{EvoEdit} method for the lifelong free-text editing task. \textsc{EvoEdit} introduces perturbation for feature augmentation, then selectively fuses parameters to balance the integration of new knowledge with the retention of previous knowledge. Experimental results demonstrate that \textsc{EvoEdit} consistently outperforms existing baselines in both new knowledge acquisition and old knowledge preservation\footnote{Code and data is available at \url{https://github.com/zeaoji/EvoEdit}.}.
\end{itemize}

\section{Related Work}
In this section, we briefly review two related topics: knowledge editing ($\S$\ref{KE}) and model merging ($\S$\ref{MM}).

\subsection{Knowledge Editing}  \label{KE}
Knowledge editing has become a rapidly advancing and active area of research, which aims to modify specific knowledge in LLMs while preserving unrelated information. Existing approaches to knowledge editing can generally be classified into three main paradigms: \textit{Locate-then-Edit}, \textit{Meta-Learning}, and \textit{Memory-based Methods}.

\textbf{Locate-then-Edit:} The locate-then-edit paradigm, exemplified by ROME \cite{rome}, r-ROME \cite{r-ROME}, MEMIT \cite{memit}, and AlphaEdit \cite{alphaedit}, employs causal tracing to precisely localize knowledge storage locations and directly modifies model parameters for efficient knowledge updates. While these methods demonstrate significant advantages in editing structured triplets, they face limitations such as over-reliance on entity localization and poor adaptability to unstructured knowledge. To address these constraints, AnyEdit \cite{anyedit} decomposes long-form knowledge into sequential chunks and iteratively edits the key token in each chunk. DEM \cite{dem} achieves distributed modification of commonsense knowledge by localizing the target editing layer. FiNE \cite{fine} and MEMLA \cite{memla} improve editing performance by refining update granularity at the neuron level. WilKE \cite{wilke} selects target layer based on the pattern matching degree of editing knowledge across different layers in LLMs. 

\textbf{Meta-Learning:} Meta-learning methods train a hypernetwork to obtain parameter modifications for editing LLMs \cite{ke,mend}. For example, KE \cite{ke} leverages a bidirectional-LSTM hypernetwork to predict parameter updates for factual edits. MEND \cite{mend} innovatively utilizes low-rank gradient decomposition to efficiently map decomposed gradients to parameter updates via a hypernetwork, significantly enhancing editing efficiency. MALMEN \cite{malman} handles multi-edit scenarios by generating weight shifts through hypernetworks and optimizing their aggregation via least-squares solutions. These methods aim to utilize meta-learning techniques to learn the optimal parameter updates, but require additional data for training the hypernetwork.

\textbf{Memory-Based:} This kind of methods explicitly stores edit examples in a memory and retrieve the most relevant edit facts to guide the model for accurate output, which leaves the model parameters unmodified \cite{serac,mello,wise}. For instance, SERAC \cite{serac} employs a classifier to determine whether the input matches the edits in memory. If the input matches any edit in memory, an auxiliary trained model makes predictions based on the matched edit. Otherwise, the prediction of the original model is used as the output. IKE \cite{ike} adopts the in-context learning approach for knowledge editing by using target edits as demonstrations. GRACE \cite{grace} retrieves discrete codebook entries to overwrite intermediate activations during inference to enable precise and local edits.

While these methods have made remarkable progress, they are still mostly restricted to triplet-based and single-step editing, making it challenging to handle the more realistic task of lifelong free-text knowledge editing.

\subsection{Model Merging}  \label{MM}
Model merging has emerged as an effective technique for integrating the parameters of multiple independently trained models, each encapsulating distinct knowledge or capabilities, to construct a more general and versatile model. Existing approaches to model merging can generally be divided into two categories: \textit{Weight-based Merging Methods} and \textit{Subspace-based Merging Methods}.

\textbf{Weight-based Merging Methods:} The most straightforward form of model merging involves directly averaging the parameters of multiple models. To enhance performance beyond this simple strategy, several advanced methods have been proposed to more effectively model parameter interactions \cite{akiba2025evolutionary, liu2024checkpoint, RegMean}. For instance, Task Arithmetic \cite{ilharcoediting} introduces the concept of ``task vector'', enabling manipulation of LLMs behavior through vector operations. AdaMerging \cite{yangadamerging} employs gradient descent to learn optimal merging coefficients by minimizing entropy as a surrogate loss on unlabeled test data. Fisher-Merging \cite{matena2022merging} utilizes the diagonal of each model’s Fisher information matrix to guide the merging process. MetaGPT \cite{zhou2024metagpt} frames model merging as a multi-task learning problem, aiming to minimize the average loss discrepancy between the merged model and its constituent task-specific models.

\textbf{Subspace-based Merging Methods:} These methods aim to project model parameters into a lower-dimensional subspace before performing merging operations \cite{yadav2023resolving,tang2023concrete,wang2024localizing}. For instance, TIES-Merging \cite{yadav2023resolving} first prunes task vectors by retaining only the most influential parameter values, followed by resolving sign inconsistencies among the merged models. Similarly, DARE \cite{yu2024language} removes redundant parameters within each model prior to merging, thereby reducing potential interference among multiple models. Model Tailor \cite{zhu2024model} identifies a sparse mask to capture the most impactful parameters for adaptation, and subsequently introduces a weight compensation mechanism to improve performance on both target and original tasks. EMR-Merging \cite{huang2024emr} first selects a unified model from the set of model weights, then generates lightweight, task-specific modulators such as masks and rescalers, to align the directional and magnitude differences between the unified model and each individual model.

These methods have demonstrated success in model merging, however, their applicability to the continual learning of LLMs and their potential to mitigate catastrophic forgetting remain to be investigated.

\section{Task Definition}
We propose the Lifelong Free-text Knowledge Editing (LF-Edit) task, which enables LLMs to continuously learn knowledge from free-text inputs, and preserves previous knowledge at the same time. Formally, LLMs can be formalized as parametric mapping functions: $f_{\bm{\theta}}: \mathcal{X} \rightarrow \mathcal{Y}$, where $\bm{\theta}$ denotes the parameters of the LLM. It maps the input $x\in \mathcal{X}$ to an output prediction $f_{\bm{\theta}}(x)\in \mathcal{Y}$. The LF-Edit task requires LLMs to continuously receive a free-text stream of edit instances $S=\{S_1, S_2, \ldots, S_n\}$, where the instance $S_{i}=(x_{e}^{i}, y_{e}^{i})$ represents the $i$-th free-text edit instance in the sequence. LLMs perform dynamic updates through continual knowledge editing, where each update builds upon the model state obtained from the preceding edit. This iterative process can be formally expressed as follows:
\begin{equation}
\label{LFKE}
f_{\bm{\theta}_i} = \text{LF-Edit}(f_{\bm{\theta}_{i-1}}, S_i),
\end{equation}
where $f_{\bm{\theta}_{i}}$ denotes the model after the $i$-th knowledge editing. For the $i$-th editing update, it is considered for successful when the following criteria are satisfied:
\begin{equation}
\label{KE_result}
f_{\bm{\theta}_{i}}(x_{e}^{i}) =
\begin{cases}
y_{e}^{i} & \text{if } x_{e}^{i} \in I(x_{e}^{i}, y_{e}^{i}) \\
f_{\bm{\theta}_{i-1}}(x_{e}^{i}) & \text{if } x_{e}^{i} \in O(x_{e}^{i}, y_{e}^{i}),
\end{cases}
\end{equation}
where $f_{\bm{\theta}_{i-1}}$ denotes the edited model from the previous step. $(x_{e}^{i}, y_{e}^{i})$ is the target editing instance, with the condition that $f_{\bm{\theta}_{i-1}}(x_{e}^{i}) \neq y_{e}^{i}$. The collection of such instances is referred to as the editing scope $I(x_{e}^{i}, y_{e}^{i})$, whereas the complementary set, $O(x_{e}^{i}, y_{e}^{i})$, contains inputs unrelated to the editing instances. The first criterion measures \textbf{Efficacy} that assesses whether the model successfully internalizes the new knowledge. The second criterion is \textbf{Specificity}, which evaluates whether the editing procedure preserves the integrity of previously acquired or edited knowledge, ensuring that unintended interference does not occur. In real-world applications, new knowledge continuously emerges in the form of unstructured text. To ensure that LLMs remain up to date, it is essential to perform continual knowledge editing, enabling them to assimilate new information while retaining previously acquired knowledge. However, the task poses significant challenges, as it demands high-quality datasets and effective methodologies to balance learning and retention.

% Lifelong free-text knowledge editing must satisfy dual constraints: for inputs $x \in T$, the model should produce the newly injected knowledge; for other inputs $x \in O$, where $O$ represents the non-edited input set, the original prediction should be preserved. Formally, this can be written as:
% \begin{equation}
% \label{KE_result}
% y =
% \begin{cases}
% f_{\theta_{edit}}(x) & \text{if } x \in T \\
% f_{\theta_{orig}}(x) & \text{if } x \in O
% \end{cases}
% \end{equation}
% where $f_{\theta_{edit}}$ denotes the edited model, $f_{\theta_{edit}}(x)$ is the desired output reflecting the injected knowledge, and $f_{\theta_{orig}}(x)$ is the original output prior to editing.

\begin{figure*}[t]
    \centering
    \includegraphics[width=0.8\textwidth]{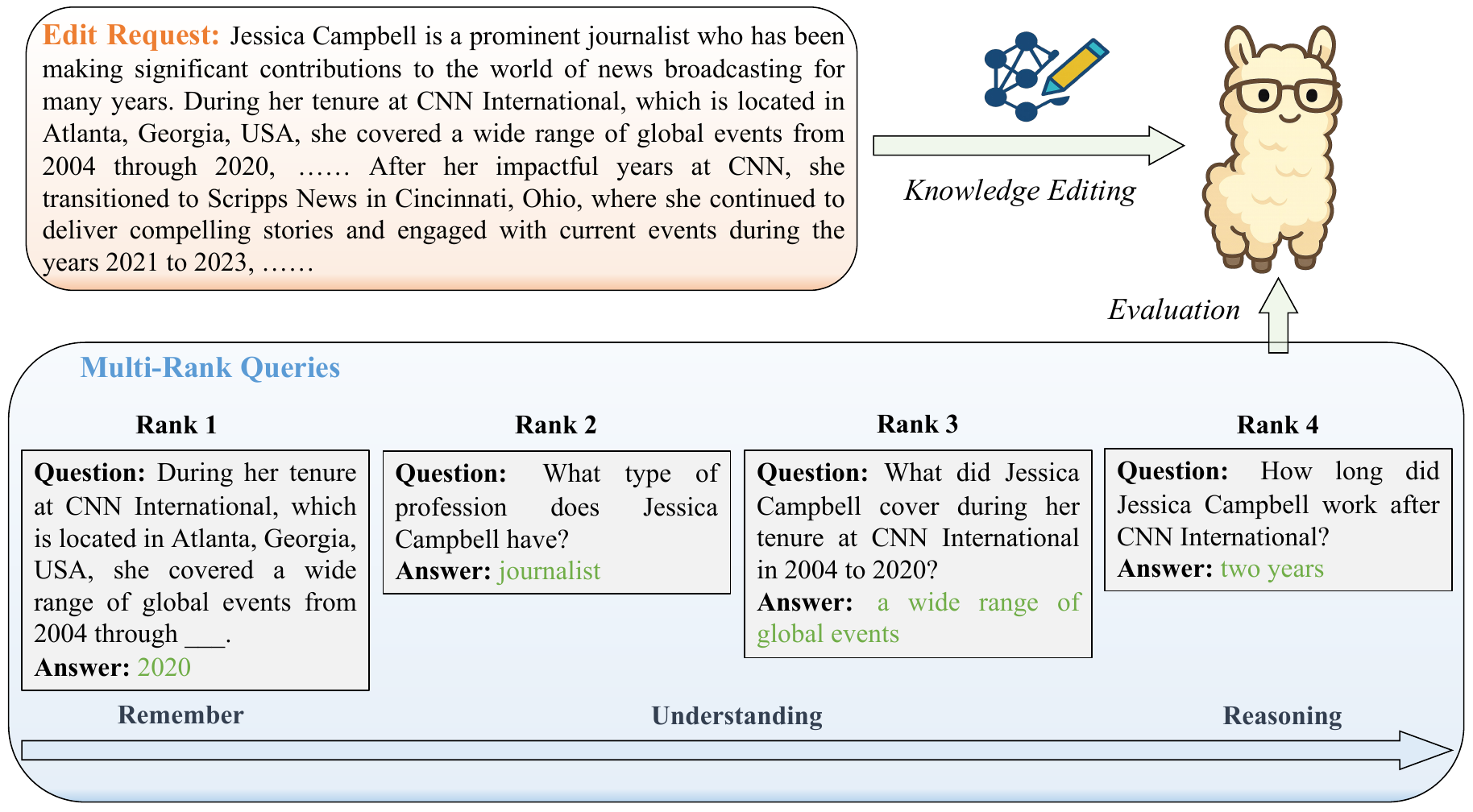}
    \caption{An example of a free-text knowledge editing instance. The edit request serves to update the model, while multi-rank queries are employed to assess the performance of the edited model.}
    \label{fig:dataset}
\end{figure*}

\section{MRLF-Bench Construction}
To advance research on the new task, we introduce a large-scale benchmark named MRLF-Bench to evaluate the effectiveness of editing over time. As shown in Figure \ref{fig:dataset}, an editing instance includes the \textit{edit request} and \textit{multi-rank queries}. In this section, we will first introduce the multi-rank evaluation designs ($\S$\ref{mrefficacy}) and data collection procedure ($\S$\ref{collection}), then summarize the benchmark statistics ($\S$\ref{data_statistics}).

\subsection{Multi-Rank Evaluation} \label{mrefficacy}
Intuitively, an effective edit should yield not only accurate recall of the original content but also robust generalization across a range of related queries. Accordingly, our evaluation framework is grounded in Piaget’s theory of cognitive development \cite{Piaget}, which posits that human cognition advances through qualitatively distinct stages. Inspired by this perspective, we construct a multi-level evaluation framework that spans a continuum from superficial memorization to higher-order generalization and reasoning:
\begin{itemize}
\item \textbf{Rank 1 (Memory Recall Level):} The questions at this level are cloze-type items that require completing missing segments from the original text. The edited model is expected to retain and reproduce the surface form of newly introduced information, enabling it to accurately fill in partial content.

\item \textbf{Rank 2 (Basic Comprehension Level):} The questions at this level involve simple synonymous rephrasings or paraphrases of the source text. The edited model should be capable of understanding and responding correctly to semantically equivalent expressions. 

\item \textbf{Rank 3 (Constrained Comprehension Level):} The questions at this level incorporate specific constraints or conditions that test fundamental understanding. The edited model should grasp these constraints in full to produce precise and contextually appropriate predictions. For instance, as illustrated in Figure \ref{fig:dataset}, the edited model needs to account for temporal constraints to answer the query correctly.

\item \textbf{Rank 4 (Complex Reasoning Level):} The questions at this level require higher-order reasoning and the integration of multiple pieces of information that go beyond literal understanding. The edited model should demonstrate deep conceptual understanding and apply reasoning based on diverse textual cues derived from the newly incorporated knowledge. For example, to answer the query of Rank 4 in Figure \ref{fig:dataset}, the edited model must first infer that the next workplace is Sripps News, then determine the time from 2021 to 2023, and finally deduce that the duration is two years.
\end{itemize}

Based on a multi-rank evaluation framework, we can systematically evaluate the edited model, which can promote the development of the lifelong free-text editing task.

\subsection{Data Collection and Curation} \label{collection}
The dataset is primarily derived from Wikidata\footnote{\url{https://www.wikidata.org/wiki/Wikidata:Main_Page}}, which was specifically chosen due to its extensive collection of entities rich in both temporal and spatial attributes. The data spans multiple domains, including sports, media, education, and politics. These domains are particularly suitable for our study, as they undergo frequent and significant knowledge updates, making them ideal for constructing data for the LF-Edit task. Wikipedia typically provides abundant structured information for each entity in the form of relational triples. To adapt this information to our purposes, we convert the triples into free-text form through the following steps:

\textbf{1) Sentence Generation:} We utilize GPT-4o-mini, guided by meticulously crafted prompts, to transform structured triples into fluent and informative sentences, each of which incorporates multiple entities and relationships.
    
\textbf{2) Counterfactual Rewriting:} To simulate the acquisition of new knowledge, we apply counterfactual rewriting to the generated texts, inspired by the approach in the CounterFact dataset \cite{rome}. Specifically, we employ LLMs to revise the original text into counterfactuals that are not encountered during pretraining. Following rewriting, we implement quality control by filtering out low-quality samples and regenerating them to ensure coherence and factual accuracy. The final dataset consists of 16,835 counterfactual free-text instances. 
    
\textbf{3) Question Generation:} After constructing the texts, we leverage GPT-4o-mini to generate questions across various domains, with a particular focus on queries designed to evaluate a model's ability to comprehend, retain, and reason about the newly acquired knowledge. After that, we manually curate the evaluation data to ensure the quality of questions.

\begin{table}
\centering
\caption{The statistical information of the MRLF-Bench. ``Edit.Len'', ``Query.Len'' and ``Ans.Len'' denote the average length of the edits, queries and corresponding answers. ``Avg.Rank'' represents the average of all rank statistics.}
\label{dataset}
\begin{tabular}{lccccc}
\toprule
\textbf{Rank}     & \textbf{\#Edits} & \textbf{Edit.Len} & \textbf{\#Query} & \textbf{Query.Len} & \textbf{Ans.Len} \\
\midrule
Rank 1   & 16,835  & 141.9    & 33,670  & 40.2      & 1.5        \\
Rank 2   & 16,835  & 141.9    & 33,670  & 10.1      & 3.6        \\
Rank 3   & 16,835  & 141.9    & 33,670  & 10.9      & 3.6        \\
Rank 4   & 16,835  & 141.9    & 33,670  & 10.7      & 4.5        \\
Avg.Rank & 16,835  & 141.9    & 33,670  & 18.0      & 3.3     \\
\bottomrule
\end{tabular}
\end{table}

\subsection{Dataset Statistics}  \label{data_statistics}
The statistical characteristics of the benchmark are summarized in Table \ref{dataset}. MRLF-Bench consists of 16,835 edit requests, with an average length exceeding 141 words. Each rank setting includes 33,670 questions, with a mean length of more than 10 words per question. The statistics further indicate that the average length of the answer across the dataset is greater than 3 words, suggesting that the responses primarily consist of phrases or concise sentences. The large scale of the MRLF-Bench enables a comprehensive and systematic evaluation of lifelong editing performance on free-text inputs across various editing algorithms.

\begin{figure*}[t]
    \centering
    \includegraphics[width=0.99\textwidth]{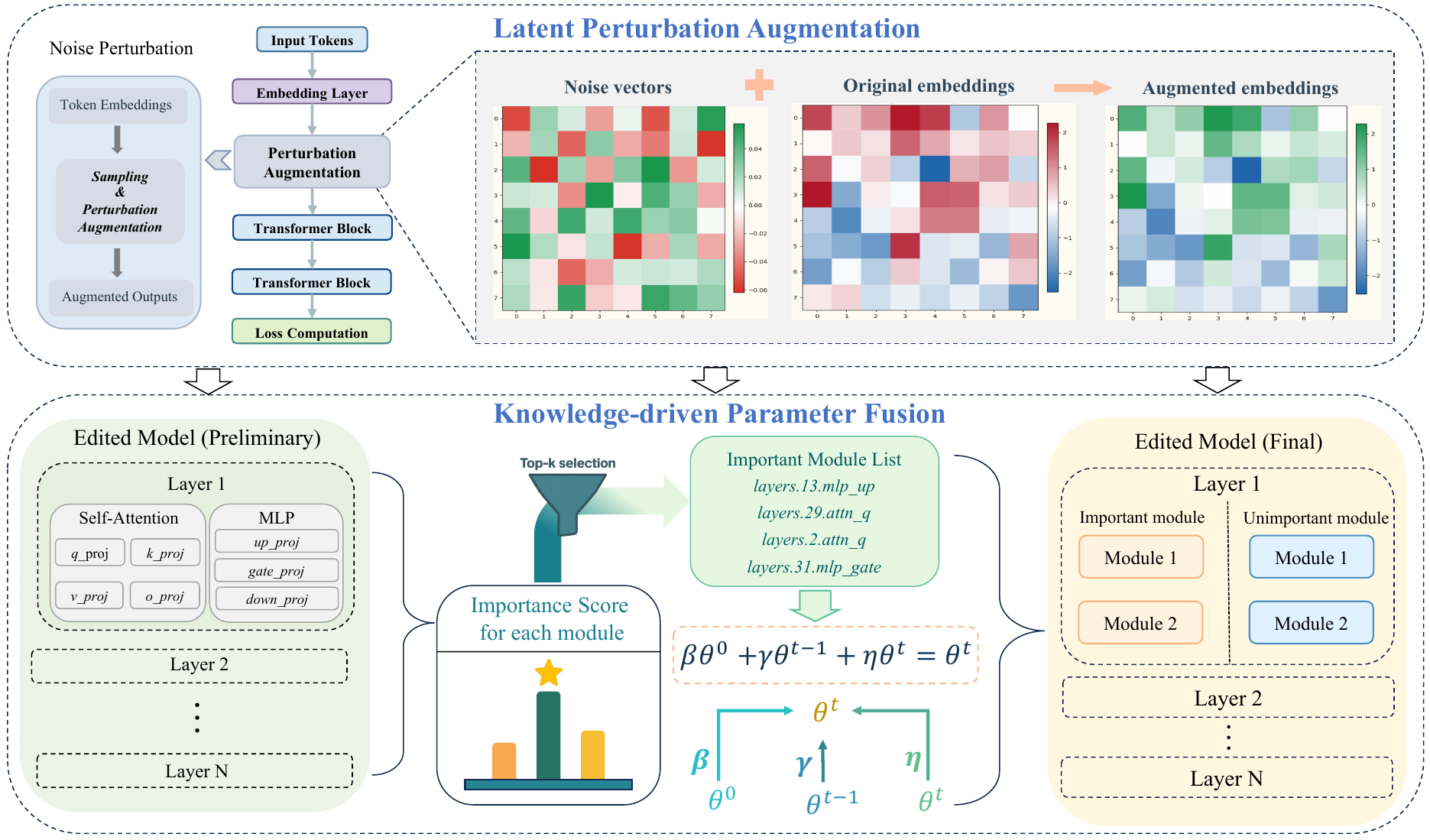}
    \caption{The architecture of the proposed \textsc{EvoEdit} for the lifelong free-text editing task, which consists of two key modules: 1) \textit{Latent Perturbation Augmentation} ($\S$\ref{PA}), which injects the noise into the embedding layer to enhance the diversity of input representations, thereby improving the model’s capacity to comprehend and assimilate new knowledge. 2) \textit{Knowledge-driven Parameter Fusion} ($\S$\ref{KPF}), which identifies and selects the most critical parameters across multiple models for integration, enabling the preservation of previously acquired knowledge.}
    \label{fig:EvoEdit}
\end{figure*}

\section{Methodology}
Figure \ref{fig:EvoEdit} presents the overall architecture of the proposed \textsc{EvoEdit}, which comprises two major components: 1) \textit{Latent Perturbation Augmentation} ($\S$\ref{PA}), designed to enhance intermediate representations and facilitate more effective knowledge acquisition in LLMs. 2) \textit{Knowledge-driven Parameter Fusion} ($\S$\ref{KPF}), which employs a parameter fusion strategy to preserve both prior and newly introduced knowledge within the edited model. Each component is described in detail in the following sections.

\subsection{Latent Perturbation Augmentation}  \label{PA}
One of the primary challenges in the LF-Edit task lies in acquiring new knowledge from free-text inputs, which often contain diverse, unstructured entities and relations. This inherent complexity makes it difficult for models to capture all relevant information during the editing process. To mitigate this issue, we propose a Latent Perturbation Augmentation (LPA) module as a data augmentation strategy, designed to enhance the ability of LLMs to assimilate new knowledge more effectively. The core idea of this module is to introduce controlled random noise into the embedding representations of LLMs during fine-tuning, thereby generating augmented inputs for the knowledge editing process. In contrast to paraphrasing-based data augmentation approaches, LPA mitigates the substantial computational overhead of iterative paraphrase generation and addresses the restricted diversity of generated text. It promotes stronger generalization from free-text inputs and enables more efficient knowledge acquisition.

Formally, LLMs receive a stream of edit instances $S=\{S_1, S_2, \ldots, S_n\}$, where the instance $S_{i}=(x_{e}^{i}, y_{e}^{i})$ represents the $i$-th free-text edit instance. Each instance is first tokenized using the model’s tokenizer:
\begin{equation}
\label{tokenizer}
\bm{t}^{i}=[t_{1}^{i}, t_{2}^{i}, \dots, t_{L}^{i}] = \text{Tokenizer}(x_{e}^{i}),
\end{equation}
where $L$ is the length of the tokenized sequence. Then, these tokens are projected into vectors through an embedding layer:
\begin{equation}
\label{embeder}
    \bm{E}^{i} = \text{Embed}(\bm{t}^{i}) = [\bm{e}_1, \bm{e}_2, \dots, \bm{e}_L] \in \mathbb{R}^{L \times d},
\end{equation}
where $d$ is the embedding dimension. To perform perturbation augmentation, we inject random noise into each token embedding vector $\bm{e}_{i}$. The noise vector $\bm{\epsilon} \in \mathbb{R}^d$ is sampled independently from a zero-mean uniform distribution scaled by a hyperparameter $\alpha$:
\begin{equation}
\label{aug}
\tilde{\bm{e}}_i = \bm{e}_i+\bm{\epsilon},\quad \bm{\epsilon}\sim U\left(-\frac{\alpha}{\sqrt{L}d},\frac{\alpha}{\sqrt{L}d}\right).
\end{equation}
The noise augmented embedding sequence is denoted as $\tilde{\bm{E}}^{i} = [\tilde{\bm{e}}_1, \tilde{\bm{e}}_2, \ldots, \tilde{\bm{e}}_L]$, which is then passed into the rest of the model for prediction. The loss function (i.e., language modeling loss) based on the augmented edit is computed as follows:
\begin{equation}
\label{loss}
\mathcal{L}(S_{i}, \bm{\theta}) = -\sum_{l=1}^{L}\text{log}p(t_{l}\, |\, t_{<l}; \tilde{\bm{E}}^{i}),
\end{equation}
where $p(\cdot)$ represents the prediction probability of the model, and $t_{<l}$ denote the previous generated sequence.

% To improve the model's ability to answer questions, we have also introduced the following loss:
% \begin{equation}
% L_{t} = -\text{log}p(a|q),
% \end{equation}
% where $(q, a)$ is the QA pair that is constructed based on the edit. Combining the above objectives, we update the model parameters with the final loss function:
% \begin{equation}
% \label{loss}
% L_{total} = \beta L_{e}+\gamma L_{t},
% \end{equation}
% where $\beta$ and $\gamma$ are hyper-parameters that are searched on the training data. 

The LPA module operates as a fine-grained data augmentation mechanism designed to enhance the model’s ability to internalize free-text knowledge during the editing process. Concurrently, it functions as a regularization strategy that mitigates overfitting and strengthens generalization capacity. More specifically, by injecting controlled noise into token embeddings during the editing phase, the model is encouraged to assimilate the underlying semantic content rather than depend on exact token sequences or superficial formatting cues. This approach effectively reduces overfitting to specific textual patterns and promotes robust generalization to novel or rephrased inputs. As a result, the model acquires knowledge in a more flexible and abstract form, enabling adaptive reasoning across diverse query formulations rather than relying on verbatim memorization.

\subsection{Knowledge-driven Parameter Fusion}   \label{KPF}
Intuitively, the current model tends to better grasp newly introduced knowledge, whereas the original model and the previously edited model preserve earlier learned and edited knowledge, respectively. To address the challenge of catastrophic forgetting in lifelong model editing, we introduce a Knowledge-driven Parameter Fusion (KPF) module. This module operates in two stages: First, after forward computation, it estimates the contribution of each model component to the current knowledge being injected. Next, it selectively merges the most informative parameters from three sources: the original model, the previously edited model, and the newly updated model. Through this selective parameter integration, KPF effectively preserves prior knowledge while assimilating new information in a balanced and controlled manner.

\textbf{Importance Calculation:} For each layer $l$ in the model, we choose the parameter matrix $\bm{\theta}_{c}$ from the parameter matrix of Self-Attention (Attn) and Multi-layer Perceptron (MLP) to calculate the importance score:  
\begin{equation}
\begin{split}
\bm{\theta}_{c} \in \{\, &\textit{\text{attn\_q}},\ \textit{\text{attn\_k}},\ \textit{\text{attn\_v}},\ \textit{\text{attn\_o}}, \\
          &\textit{\text{mlp\_gate}},\ \textit{\text{mlp\_up}},\ \textit{\text{mlp\_down}} \,\}.
\end{split}
\end{equation}
For parameters set $\bm{\theta}_{c}$, the importance score $\mathcal{S}_c$ is computed as follows when editing the instance $S_{i}$:
\begin{equation}
\begin{aligned}
\mathcal{S}_{c}(S_{i}, \bm{\theta}) &=\left|\mathcal{L}(S_{i},\bm{\theta})-\mathcal{L}\left(S_{i},\bm{\theta}\mid\bm{\theta}_{c}=0\right)\right| \\
& = \left|\frac {\partial \mathcal{L}}{\partial \bm{\theta}_{c}} (\bm{\theta}_{c} - 0) + \frac{1}{2!}\frac{\partial^{2} \mathcal{L}}{\partial \bm{\theta}_{c}^{2}}(\bm{\theta}_{c} - 0)^{ 2 } + \cdots \right| \\
& \approx \left| \bm{\theta}_{c}^{T} \frac {\partial \mathcal{L} } { \partial \bm{\theta}_{c} } \right|. \\
\end{aligned}
\label{formula:importance_score}
\end{equation}
To accelerate the calculating speed, the score is estimated by calculating the first-order derivative of the Taylor expansion in the above formula. Accumulating these scores over all layers and components yields a global importance list: 
\begin{equation}
\mathcal{S} = [\mathcal{S}_{1,\text{attn\_q}}, \mathcal{S}_{1,\text{attn\_k}}, \ldots, \mathcal{S}_{H, \text{mlp\_up}}, \mathcal{S}_{H, \text{mlp\_down}}],
\end{equation}
where $H$ is the number of layers. $\mathcal{S}_{i,m}$ represents the importance score corresponding to parameter set $\bm{\theta}_{m}$ in $i$-th layer. 

\textbf{Parameter Merging:} We sort the importance scores in descending order, and select the top $k\%$ of components based on these scores to form the final important parameters:
\begin{equation}
\label{selection}
    \mathcal{M}_k = \{ \bm{\theta}_{m} \mid \mathcal{S}_{i,m} \text{ is among the top } k\% \}.
\end{equation}

After obtaining important parameters, the next step is to perform parameter fusion. $\bm{\theta}^0$ represents the original (pre-editing) parameters, $\bm{\theta}^{t-1}$ denotes the last step parameters, and $\bm{\theta}^{t}$ be the current updated parameters. For each component index $i$, we fuse the above three types of parameters according to the following formula:
\begin{equation}
\label{merge}
    \bm{\theta}_i^t = \begin{cases} \beta \bm{\theta}_i^0 + \gamma \bm{\theta}_i^{t - 1} + \eta \bm{\theta}_i^{t}, & \bm{\theta}_{i} \in M_k, \\ \bm{\theta}_i^{t}, & \bm{\theta}_{i} \notin M_k, \end{cases}
\end{equation}
where $\beta$, $\gamma$ and $\eta$ are fusion coefficients ($\beta + \gamma + \eta = 1$). By restricting fusion with $\mathcal{M}_k$, we ensure that historical information and newly learned knowledge, that is, the knowledge stored in important components, are well retained. Meanwhile, other knowledge of the model is preserved through unimportant components. The overall procedure of the proposed method \textsc{EvoEdit} is outlined in Algorithm \ref{alg:Framwork}.

\begin{algorithm}[t]
	\caption{The Procedure of \textsc{EvoEdit} Method} 	
	\label{alg:Framwork} 
	\begin{algorithmic}[1]
  \REQUIRE  Large language model $f$, a stream of edit instances $S=\{S_1, S_2, \ldots, S_n\}$, hyperparameters $\alpha$, $\beta$, $\gamma$ and $\eta$.  \\
  \ENSURE Optimized model parameters. \\
      \FOR {Instance $S_{i}$ in $\{S_1, S_2, \ldots, S_n\}$}
		\STATE Project tokens into vectors via Equation (\ref{tokenizer}) and Equation (\ref{embeder}) \\
        \STATE Perform perturbation augmentation for each token via Equation (\ref{aug}) \\
        \STATE Feed the augmented embedding sequence into the rest of the model for prediction \\
        \STATE Update the model parameters with the loss function via Equation (\ref{loss}) \\
        \FOR {each parameter matrix $\bm{\theta}_{c}$}
		\STATE Compute the importance score $\mathcal{S}_c$ via Equation (\ref{formula:importance_score}) \\
        \STATE Add the importance score $\mathcal{S}_c$ into the global importance list $\mathcal{S}$
		\ENDFOR
        \STATE Select the top $k\%$ of components based on the importance scores
        \STATE Merge the original parameters, the last step parameters and the current updated parameters via Equation (\ref{merge})
      \ENDFOR
	\end{algorithmic}	
\end{algorithm}

\begin{table*}
    \caption{Experimental results of different number of edits on the MRLF-Bench. The BLEU and PPL denote the average of four rank results. The ``--'' means perplexity exceeded 10,000. The best results are highlighted with \textbf{Bold}.}
    \centering
   \label{tab:main}
    \begin{tabular}{c|l|ccccccccccc}
    \toprule
    \multirow{2}{*}{\textbf{Base Model}}   & \multirow{2}{*}{\textbf{Method}} & \multicolumn{2}{c}{\textbf{T=100}}    & \multicolumn{2}{c}{\textbf{T=500}}    & \multicolumn{2}{c}{\textbf{T=1000}}    & \multicolumn{2}{c}{\textbf{T=1500}}    &  \multicolumn{2}{c}{\textbf{T=2000}}  \\
     &  & \textbf{BLEU}$\uparrow$ & \textbf{PPL}$\downarrow$ & \textbf{BLEU}$\uparrow$ & \textbf{PPL}$\downarrow$ & \textbf{BLEU}$\uparrow$ & \textbf{PPL}$\downarrow$ & \textbf{BLEU}$\uparrow$ & \textbf{PPL}$\downarrow$  & \textbf{BLEU}$\uparrow$ & \textbf{PPL}$\downarrow$  \\ 
     \midrule
    \multirow{7}{*}{\textbf{LLaMA-3}} & Pre-Editing        & 36.20  &40.77  &37.89  &25.50  &37.55  &34.48  &37.43  &24.32  &  37.50 & 24.05 \\
                                     & Fine-Tuning             &59.85   &42.70  &60.24  &34.09  &60.14  &32.95  &60.27  &31.54  & 60.42 & 30.33 \\
                                     & ROME                    &0.81   & --  &0.91  & --  &1.00  & -- &1.10  & --  &  1.14 & -- \\
                                     & MEMIT                   &36.36  &40.77  &34.37  &71.17  &25.04  &641.18  &18.83  &3299.03  & 15.08 & 7909.69 \\
                                     & AlphaEdit               &39.49   &30.92  &39.53  &31.94  &30.59  &123.36  &23.59  &545.80  & 19.09 & 1738.88 \\
                                     & MEND                    &0.83   & --  &15.05  & 6576.48  & 17.14  & 2577.66  & 16.05  & 3372.93  & 15.14 & 4106.28 \\
    \cmidrule(r){2-12}
                        &  \textbf{\textsc{EvoEdit} (Ours)}                    &\textbf{64.86}   &\textbf{7.22}  &\textbf{65.61}  &\textbf{6.23}  &\textbf{64.92}  &\textbf{6.56}  &\textbf{64.84}  &\textbf{6.61}  & \textbf{64.85} & \textbf{6.52} \\
    \midrule
    \midrule
    \multirow{7}{*}{\textbf{LLaMA-2}} & Pre-Editing      &35.94   &16.65  &35.94  &14.82  &35.67  &14.31  &35.43  &14.37  &  35.56 & 14.40 \\
                                     & Fine-Tuning             &66.64   &5.90  &66.82  &5.26  &65.28  &5.42  &64.48  &5.55  & 64.11 & 5.59 \\
                                     & ROME                    &0.68   & --   &0.87  & --  &0.91  & --  &0.93  & --  & 0.94 & -- \\
                                     & MEMIT                   &7.47   &--  &2.51  & --  &1.59  & --  &1.24 & --  & 1.03 & -- \\
                                     & AlphaEdit               &22.40   &8.49  &12.00  &878.68  &7.45  &2686.11  &5.57  &4688.85  & 4.59 & 6422.04 \\
                                     & MEND                    &17.74   &247.49  &27.17  &64.56  &28.82  &45.91  &29.05  &54.06  & 28.99 & 41.46 \\
        \cmidrule(r){2-12}                             
                                     &  \textbf{\textsc{EvoEdit} (Ours)}                   &\textbf{71.19}   &\textbf{3.82}  &\textbf{70.96}  &\textbf{3.65}  &\textbf{70.24}  &\textbf{3.70}  &\textbf{69.91}  &\textbf{3.75}  &  \textbf{69.90} & \textbf{3.75} \\ 
    \midrule
    \midrule
    \multirow{7}{*}{\textbf{GPT-J}}   & Pre-Editing              &29.92   &32.88  &29.91  &30.64  &29.63  &29.22  &29.54  &28.84  & 29.50 & 28.63 \\
                                     & Fine-Tuning             &54.76   &13.47  &53.65  &12.23  & 53.14  &12.34  &53.08  &12.49  & 53.11 & 12.47 \\
                                     & ROME                    &2.01   & --  &3.27  & --  &3.43  & --  &3.28  & --  & 3.12 & -- \\
                                     & MEMIT                   &33.73   &26.62  &33.22  &28.12  &31.01  &34.63  &27.93  &55.02  & 24.65 & 107.25 \\
                                     & AlphaEdit               &31.69   &28.08  &27.13  &60.10  &19.59  &361.11  &15.47  &1192.35  & 12.94 & 2906.73 \\
                                     & MEND                    & 0.62   & --  &9.55  & --  &8.85  & --  &7.52  & --  & 6.77 & -- \\
        \cmidrule(r){2-12} 
                                     &  \textbf{\textsc{EvoEdit} (Ours)}                    &\textbf{62.79}   &\textbf{5.78}  &\textbf{62.03}  &\textbf{6.14}  &\textbf{61.44}  &\textbf{6.15}  &\textbf{61.10}  &\textbf{6.23}  & \textbf{60.87} & \textbf{6.22} \\ 
    \midrule
    \midrule
    \multirow{7}{*}{\textbf{GPT-2}}   & Pre-Editing              &20.40   &64.00  &19.69  &63.72  &19.71  &60.53  &19.66  &59.89  & 19.64 & 60.07 \\
                                     & Fine-Tuning             &\textbf{56.71}   & 17.57  &\textbf{56.62}  &17.70  &\textbf{56.50}  &16.13  &\textbf{56.45}  &15.60  & \textbf{56.33} & 15.23 \\
                                     & ROME                    &7.89   &1958.98  &5.42  &3860.13  &4.38  &5782.95  &3.83  &7371.26  & 3.47 & 8768.66 \\
                                     & MEMIT                   &22.44   &63.54  &21.38  &75.86  &20.08  &92.96  &18.50  &141.95  & 16.99 & 215.53 \\
                                     & AlphaEdit               &22.52   &71.33  &17.94  &177.79  &13.51  &645.10  &10.78  &1400.43  & 9.07 & 2267.90 \\ 
                                     & MEND                    &19.09   &107.82  &19.41  &93.38  &19.66  &65.46  &19.75  &63.63  & 19.78 & 62.97  \\
        \cmidrule(r){2-12}                   
                                     &  \textbf{\textsc{EvoEdit} (Ours)}                    &49.59   &\textbf{12.28}  &48.13  &\textbf{12.09}  &47.51  &\textbf{11.81}  &47.53  &\textbf{11.88}  &  47.53 & \textbf{12.03} \\ 
        \bottomrule
    \end{tabular}
\end{table*}

\section{Experiments}
In this section, we conduct extensive experiments with the aim of answering the following research questions (\textbf{RQs}):
\begin{itemize}
  \item \textbf{RQ1}: How effective is our method \textsc{EvoEdit} in learning new knowledge? ($\S$\ref{new_results})
  \item \textbf{RQ2}: How effective is our method \textsc{EvoEdit} in preserving previous knowledge? ($\S$\ref{old_knowledge})
  \item \textbf{RQ3}: How does each design of the proposed method matter? ($\S$\ref{AS})
\end{itemize}

In the remainder of this section, we describe the knowledge editing baselines ($\S$\ref{KEB}), the adaptation for existing editing methods ($\S$\ref{adaptation}), evaluation metrics ($\S$\ref{EM}), and the case study ($\S$\ref{case study}).

\subsection{Knowledge Editing Baselines}  \label{KEB}
To evaluate the effectiveness of \textsc{EvoEdit}, we compare it with four distinct categories of editing approaches:

\textbf{1) No-Edit:} To establish a baseline, we evaluate the performance of the original model without any editing techniques, referred to as \textit{Pre-Editing}. This setting reflects the model’s initial capability in answering queries prior to incorporating knowledge edits. The resulting performance serves as a lower bound for evaluating the effectiveness of the knowledge editing methods.

\textbf{2) Fine-Tuning:} Following prior work \cite{mulfe,grace}, we perform full-parameter updates using the editing samples and select the best results across various hyperparameter configurations.

\textbf{3) Meta-Learning:} We evaluate \textit{MEND} \cite{mend}, a meta-learning-based approach that utilizes a hypernetwork to generate parameter updates, guided by edit-specific gradients. In accordance with standard practices, we train a hypernetwork on both the ZSRE \cite{zsre} and CounterFact \cite{rome} datasets, selecting the best-performing checkpoint for evaluation.
    
\textbf{4) Locate-then-Edit:} We select \textit{ROME} \cite{rome}, \textit{MEMIT} \cite{memit}, and \textit{AlphaEdit} \cite{alphaedit} as representative methods within the locate-then-edit paradigm. These methods utilize causal tracing to identify parameters associated with the target knowledge and perform direct updates to those parameters. MEMIT extends ROME by enabling batch editing, while AlphaEdit enhances specificity by projecting perturbations onto the null space of the preserved knowledge before applying them to the parameters. Since these methods are not originally designed for free-text inputs, we apply triple extraction to facilitate a fair comparison, as detailed in Section \ref{adaptation}.

\subsection{Triple Extraction for Knowledge Editing Methods}  \label{adaptation}
Most existing knowledge editing methods, such as ROME, MEMIT, and AlphaEdit, are originally designed to operate on structured triplets. To extend these methods to lifelong knowledge editing in a free-text setting, we convert unstructured edit requests into structured triplets through a simplification strategy. Specifically, we employ Open Information Extraction (OpenIE), which extracts relational triplets from natural language text without relying on predefined schemas. This enables free-text edit requests to be transformed into structured representations compatible with established editing approaches. In our implementation, we adopt the Stanford OpenIE toolkit \cite{stanford-openie} to extract triplets from each edit request and subsequently reformat them to conform to the input specifications of classical editing methods.

\begin{table*}
    \caption{Experimental results of the last step editing on the MRLF-Bench. The best results are highlighted with \textbf{Bold}. The ``--'' means perplexity exceeded 10,000.}
    \centering
   \label{tab1}
    \begin{tabular}{c|l|ccccccccccc}
    \toprule
    \multirow{2}{*}{\textbf{Base Model}}   & \multirow{2}{*}{\textbf{Method}} & \multicolumn{2}{c}{\textbf{Rank 1}}    & \multicolumn{2}{c}{\textbf{Rank 2}}    & \multicolumn{2}{c}{\textbf{Rank 3}}    & \multicolumn{2}{c}{\textbf{Rank 4}}    &  \multicolumn{2}{c}{\textbf{Average}}  \\
     &  & \textbf{BLEU}$\uparrow$ & \textbf{PPL}$\downarrow$ & \textbf{BLEU}$\uparrow$ & \textbf{PPL}$\downarrow$ & \textbf{BLEU}$\uparrow$ & \textbf{PPL}$\downarrow$ & \textbf{BLEU}$\uparrow$ & \textbf{PPL}$\downarrow$  & \textbf{BLEU}$\uparrow$ & \textbf{PPL}$\downarrow$  \\ 
     \midrule
    \multirow{7}{*}{\textbf{LLaMA-3}} & Pre-Editing        & 61.89  &4.73  &26.29  &35.77  &33.59  &17.46  &28.38  &37.77  &  37.54 & 23.93 \\
                                     & Fine-Tuning             &85.93   &2.31  &50.17  &44.03  &54.89  &22.19  &50.91  &50.66  & 60.48 & 29.80 \\
                                     & ROME                    &1.58   & --  &0.99  & --  &1.04  & -- &1.03  & --  &  1.16 & -- \\
                                     & MEMIT                   &22.54  &1528.60  &8.24  &--  &10.65  &--  &9.16  &--  & 12.65 & -- \\
                                     & AlphaEdit               &25.51   &636.16  &11.27  &6928.25  &14.73  &4139.61  &12.74  &4762.07  & 16.06 & 4116.52 \\
                                     & MEND                    &22.26   & 1068.21  &10.58  & 6308.49  & 14.05  & 3927.41  & 10.71  & 6969.30  & 14.40 & 4568.35 \\
    \cmidrule(r){2-12}
                        &  \textbf{\textsc{EvoEdit} (Ours)}                    &\textbf{86.46}   &\textbf{1.86}  &\textbf{56.08}  &\textbf{8.55}  &\textbf{60.58}  &\textbf{6.26}  &\textbf{56.44}  &\textbf{9.26}  & \textbf{64.89} & \textbf{6.48} \\
    \midrule
    \midrule
    \multirow{7}{*}{\textbf{LLaMA-2}} & Pre-Editing      &41.47   &4.44  &30.84  &20.06  &36.92  &11.58  &32.08  &21.54  &  35.33 & 14.41 \\
                                     & Fine-Tuning             &76.43   &1.98  &58.66  &6.91  &62.84  &4.69  &57.59  &8.86  & 63.88 & 5.61 \\
                                     & ROME                    &0.91   & --   &1.04  & --  &0.84  & --  &0.99  & --  & 0.95 & -- \\
                                     & MEMIT                   &1.78   &--  &0.56  & --  &0.70  & --  &0.55 & --  & 0.90 & -- \\
                                     & AlphaEdit               &4.76   &2791.10  &3.55  &--  &3.81  &6464.33  &3.85  &--  & 3.99 & -- \\
                                     & MEND                    &34.73   &11.60  &25.31  &52.97  &30.53  &31.83  &25.20  &66.81  & 28.94 & 40.80 \\
        \cmidrule(r){2-12}                             
                                     &  \textbf{\textsc{EvoEdit} (Ours)}                   &\textbf{84.33}   &\textbf{1.58}  &\textbf{63.71}  &\textbf{4.73}  &\textbf{68.14}  &\textbf{3.50}  &\textbf{63.50}  &\textbf{5.15}  &  \textbf{69.92} & \textbf{3.74} \\ 
    \midrule
    \midrule
    \multirow{7}{*}{\textbf{GPT-J}}   & Pre-Editing              &43.30   &16.14  &22.52  &36.78  &28.75  &20.20  &23.39  &41.16  & 29.49 & 28.57 \\
                                     & Fine-Tuning             &53.60   &8.11  &51.57  &14.30  & 56.13  &8.30  &51.30  &18.97  & 53.15 & 12.42 \\
                                     & ROME                    &3.42   & --  &2.68  & --  &3.13  & --  &2.80  & --  & 3.01 & -- \\
                                     & MEMIT                   &25.20   &132.91  &17.26  &303.71  &25.34  &129.61  &19.58  &292.13  & 21.85 & 214.59 \\
                                     & AlphaEdit               &13.85   &4967.49  &8.74  &7961.20  &12.43  &3894.83  &10.03  &6099.63  & 11.26 & 5730.79 \\
                                     & MEND                    & 8.04   & --  &5.38  & --  &6.96  & --  &5.00  & --  & 6.35 & -- \\
        \cmidrule(r){2-12} 
                                     &  \textbf{\textsc{EvoEdit} (Ours)}                    &\textbf{75.53}   &\textbf{2.95}  &\textbf{53.92}  &\textbf{8.13}  &\textbf{58.76}  &\textbf{5.26}  &\textbf{55.04}  &\textbf{8.48}  & \textbf{60.81} & \textbf{6.21} \\ 
    \midrule
    \midrule
    \multirow{7}{*}{\textbf{GPT-2}}   & Pre-Editing              &19.26   &38.91  &18.10  &73.34  &21.57  &52.45  &19.46  &77.39  & 19.60 & 60.52 \\
                                     & Fine-Tuning             &\textbf{63.73}   &\textbf{5.82}  &\textbf{52.39}  &19.77  &\textbf{55.92}  &11.77  &\textbf{53.22}  &22.68  & \textbf{56.32} & 15.01 \\
                                     & ROME                    &3.77   &--  &2.64  &--  &3.39  &6214.47  &3.01  &9523.73  & 3.20 & -- \\
                                     & MEMIT                   &22.74   &74.31  &12.06  &462.77  &14.58  &282.11  &13.88  &392.46  & 15.82 & 302.91 \\
                                     & AlphaEdit               &9.81   &2021.28  &6.36  &4520.90  &8.18  &2739.90  &7.33  &3482.09  & 7.92 & 3191.04 \\ 
                                     & MEND                    &20.48   &37.45  &18.03  &76.51  &21.28  &55.46  &19.26  &81.68  & 19.76 & 62.78  \\
        \cmidrule(r){2-12}                   
                                     &  \textbf{\textsc{EvoEdit} (Ours)}                    &37.44   &15.64  &49.24  &\textbf{11.72}  &53.34  &\textbf{7.45}  &49.94  &\textbf{13.96}  &  47.49 & \textbf{12.19} \\ 
        \bottomrule
    \end{tabular}
\end{table*}

\subsection{Evaluation Metrics}  \label{EM}
For the new LF-Edit task, we assess the effectiveness of model editing from two complementary perspectives: 1) \textit{Efficacy}: This metric evaluates the model’s acquisition of new knowledge. After editing the $i$-th instance, the edited model is assessed using multi-rank queries to determine how well it has integrated the new knowledge. 2) \textit{Specificity}: This metric examines the model’s retention of previously edited knowledge. After editing $i$-th instance, we sample multi-rank queries from previous $i-1$ instances according to a predefined sampling coefficient. In our implementation, the sampling process guarantees that at least one query from each rank of the prior instances is included.

The answers in our benchmark vary in both length and form, ranging from single words or short phrases to complete sentences. Conventional evaluation metrics such as Exact Match (EM) are not well-suited for this setting. These metrics, originally designed for triple-based knowledge editing, primarily emphasize word-level accuracy. In contrast, the responses in the new LF-Edit task typically consist of multiple words, rather than an individual word. Drawing inspiration from evaluation practices in machine translation, we employ \textbf{BLEU} \cite{bleu} as our primary metric to assess the quality of model outputs with respect to reference answers. This choice enables a more comprehensive evaluation of sentence-level generation quality in a flexible and scalable manner. Consistent with prior work \cite{mulfe}, we also report Per-Token Perplexity (\textbf{PPL}) to capture the uncertainty of the model’s predictions. Intuitively, higher BLEU scores and lower PPL values correspond to better overall model performance.

\subsection{New Knowledge Learning Results} \label{new_results}
In this subsection, we start to address the research question \textbf{RQ1}. 
Table \ref{tab:main} and Table \ref{tab1} show the new knowledge learning results of our method against knowledge editing baselines. We note the following key observations throughout our experiments:

1) Our proposed method \textsc{EvoEdit} achieves strong performance on the lifelong free-text knowledge editing task and consistently outperforms existing baselines. For instance, as shown in Table \ref{tab:main}, when built upon LLaMA-3 and evaluated with 100 edits (i.e., T=100), \textsc{EvoEdit} exceeds Fine-Tuning by 8.4\% in terms of BLEU score, and outperforms the previous state-of-the-art AlphaEdit by 64.2\%. Moreover, as the number of edited instances increases, \textsc{EvoEdit} continues to achieve substantial improvements compared to baselines. These performance gains highlight the effectiveness of \textsc{EvoEdit} in continually acquiring new knowledge.

2) Across all evaluation settings, our method \textsc{EvoEdit} achieves lower perplexity (PPL) scores when generating outputs related to newly edited knowledge compared to existing knowledge editing baselines. For instance, as shown in Table \ref{tab1} for LLaMA-3, \textsc{EvoEdit} attains a PPL of 8.55 at Rank 2, which is substantially lower than that of Fine-Tuning (i.e., 44.03) and MEMIT (i.e., 6928.25). These results demonstrate that \textsc{EvoEdit} produces more coherent and predictable outputs following knowledge edits. 

3) Existing knowledge editing methods face significant challenges in adapting to the lifelong free-text knowledge editing task. Although triple extraction provides a potential bridge between unstructured free-text inputs and structured editing techniques, current approaches such as ROME, MEMIT, MEND, and AlphaEdit still achieve relatively low BLEU scores and often exhibit high PPL. This performance degradation primarily stems from information loss introduced during the simplification process.

4) Compared with the Pre-Editing baseline, several existing methods not only fail to enhance performance but even impair the model’s ability to produce correct outputs. For instance, as shown in Table \ref{tab1}, when based on LLaMA-3, the Pre-Editing approach attains average BLEU and PPL scores of 37.54 and 23.93, respectively, outperforming the strong editing method AlphaEdit, which yields 16.06 and 4116.52. This finding indicates that current knowledge editing methods are inadequate for effectively addressing the new LF-Edit task.

\begin{figure}[t]
    \centering
    \includegraphics[width=\columnwidth]{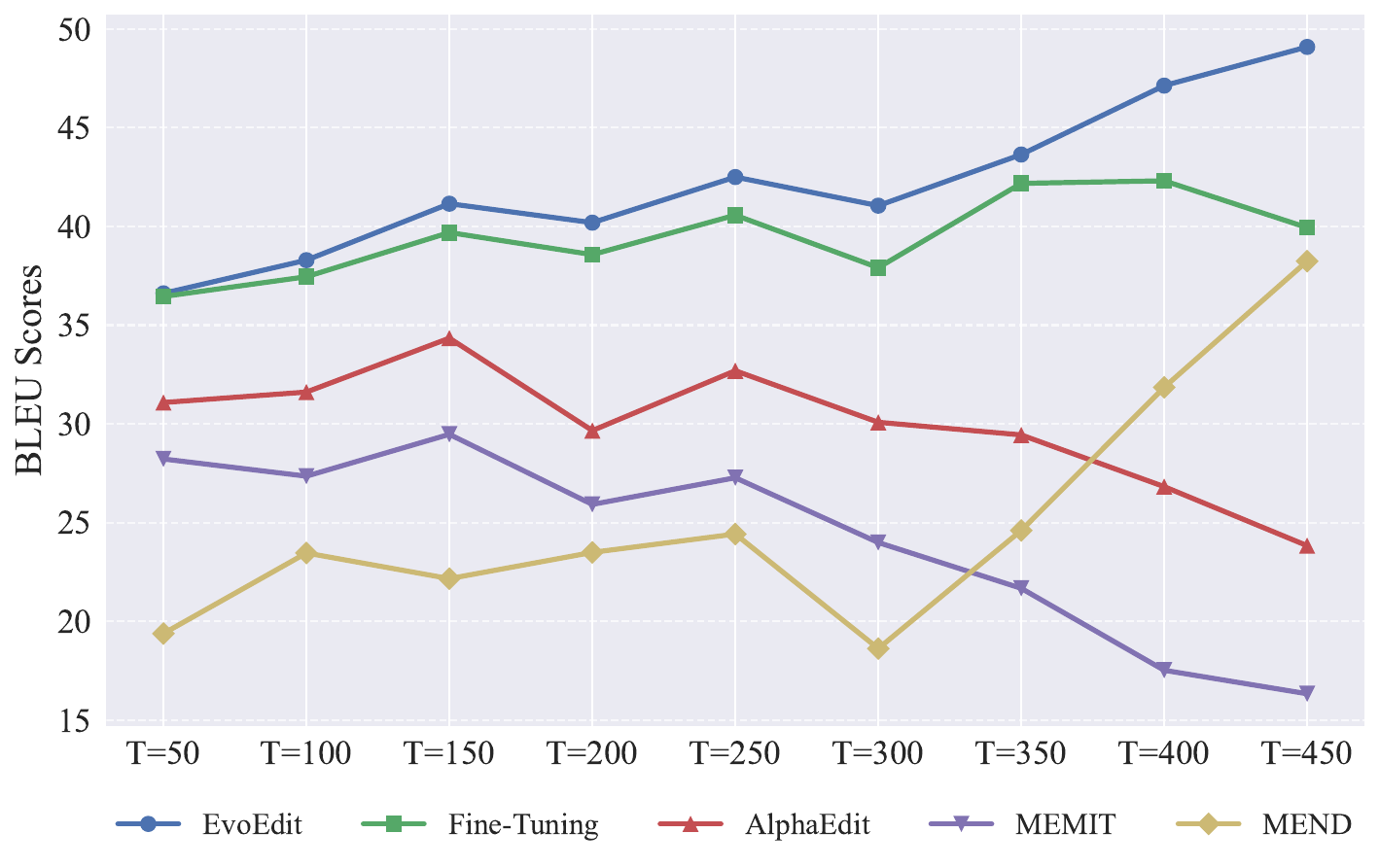} 
    \caption{BLEU scores on historical edits after after editing 500 instances for the knowledge editing methods. T=X represents the performance of the edited model on multi-rank queries at step X/50 (editing 50 instances at each step). }
    \label{fig:llama3_forgetting_step_10}
\end{figure}

\begin{figure*}[t]
	\centering         
		\includegraphics[scale=0.35]{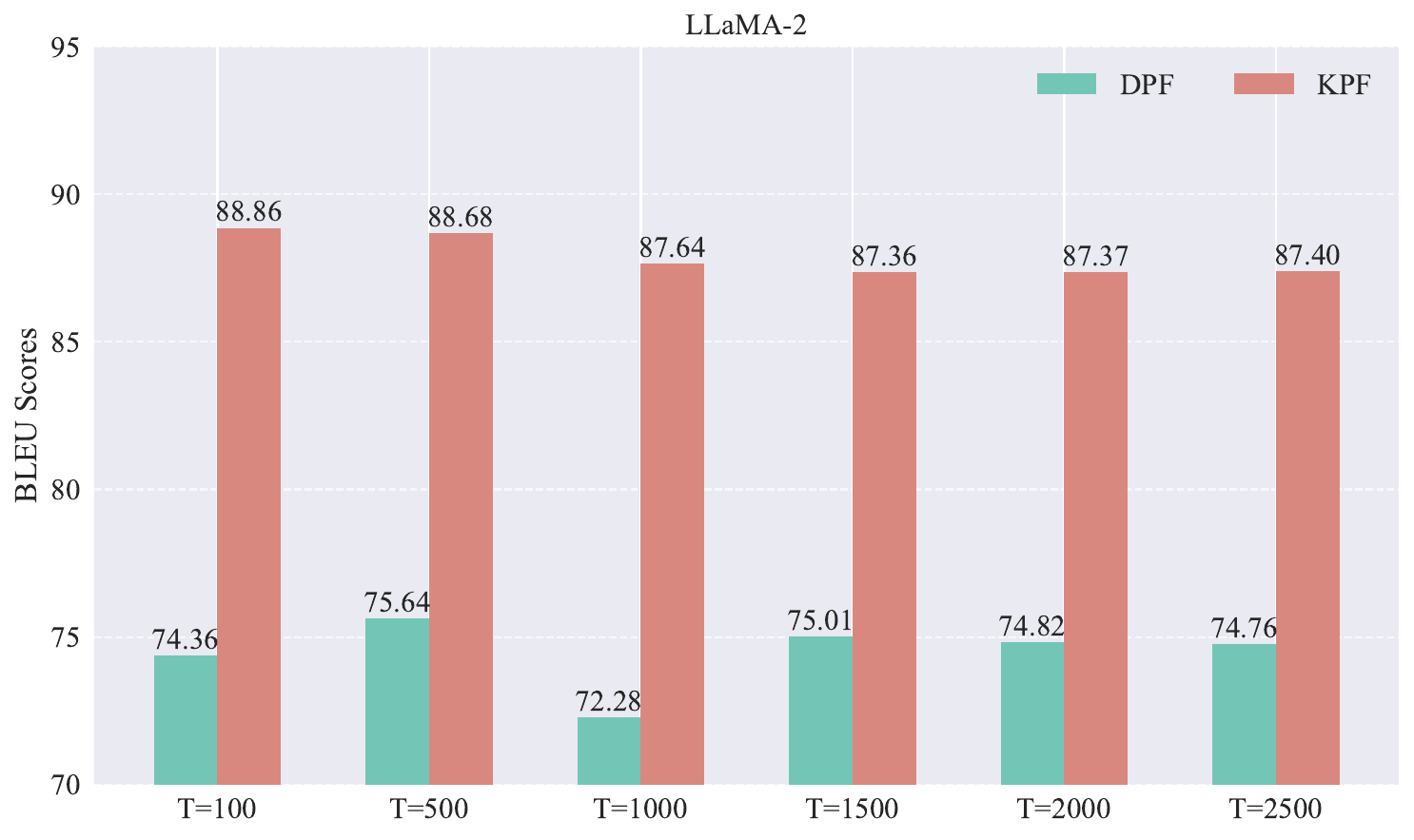}
  \hspace{0.6mm}        
		\includegraphics[scale=0.35]{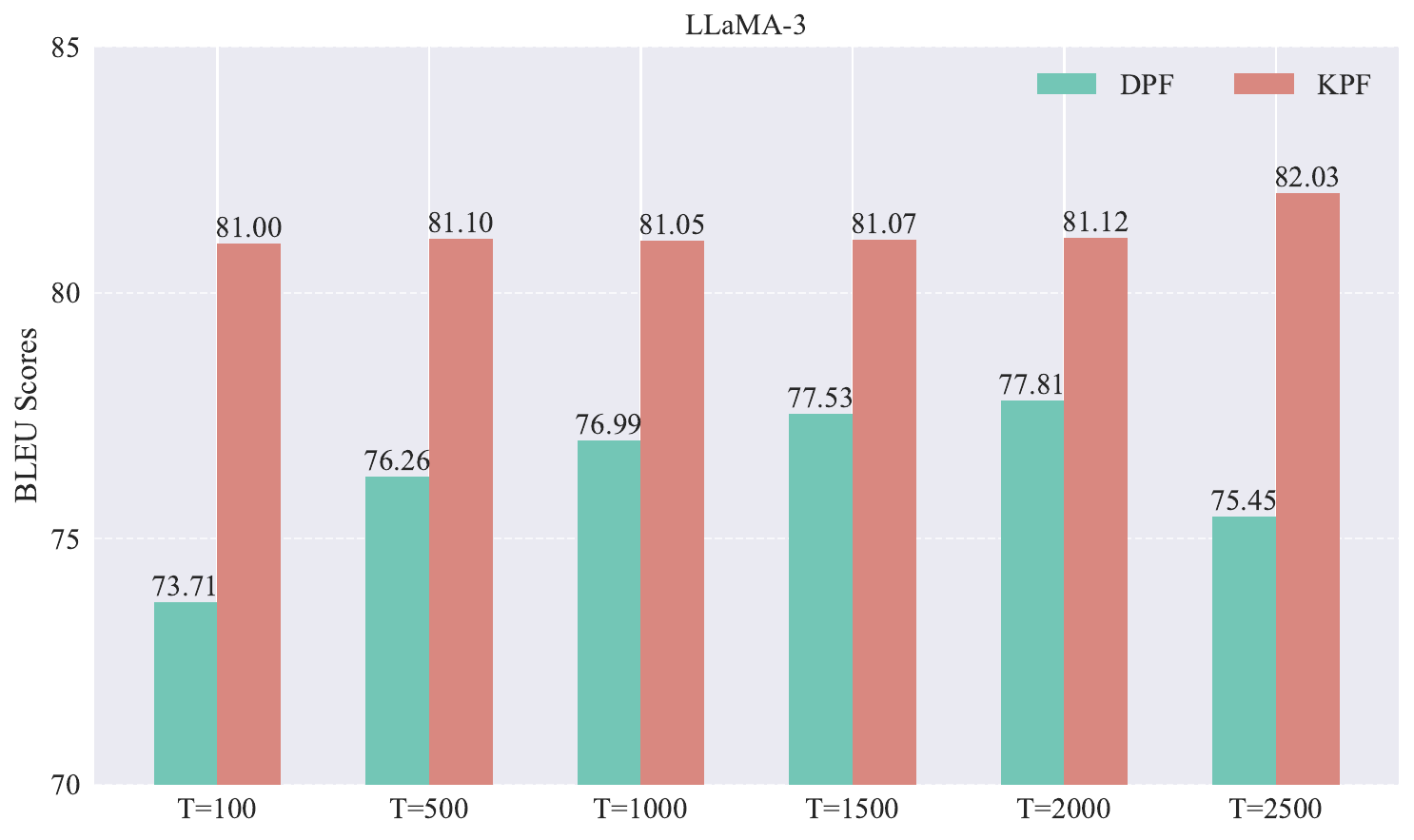}
	\caption{The performance comparison between different parameter fusion strategies on the LLaMA-2 and LLaMA-3, respectively. ``DPF'' represents the direct fusion of the original model, the edited model from the last step, and all parameters of the current model, without considering the importance of the parameters.}        
	\label{parameter-merging}                          
\end{figure*}

\subsection{Previous Knowledge Preservation Results} \label{old_knowledge}
In this subsection, we present experiments designed to evaluate the effectiveness of the proposed method in preserving previously acquired knowledge during continual editing (\textbf{RQ2}). After each editing step, the model is prompted with all questions from earlier steps, and its performance is assessed using the BLEU score. As illustrated in Figure~\ref{fig:llama3_forgetting_step_10}, upon completing the tenth editing step (i.e., T=500), we evaluate the model on a cumulative set of questions collected from prior steps. The figure demonstrates that our method consistently achieves high BLEU scores, indicating strong knowledge retention. In contrast, existing knowledge editing approaches show pronounced limitations in this aspect. For instance, methods such as MEMIT and AlphaEdit experience significant performance deterioration under sequential editing. Moreover, as the number of edits increases, the performance gap between our approach and the baselines widens. This is because continual editing makes it increasingly difficult for existing methods to integrate new knowledge without disrupting prior information. These results highlight the superior capability of our method to effectively preserve previously learned knowledge.

\begin{table}[t]
\centering
\caption{Ablation study by removing the main components with the increase of the number of edits, where ``w/o'' indicates without. The BLEU and PPL denote the average of all rank results. The best results are highlighted with \textbf{Bold}.}
\label{tab:ablation_study_on_modules}
\begin{tabular}{c|c|ccc}
\toprule
\multicolumn{1}{c|}{\textbf{Metric}}     & \multicolumn{1}{c|}{\textbf{\#Edits}} & \textbf{EvoEdit} & \textbf{w/o LPA} & \multicolumn{1}{c}{\textbf{w/o KPF}} \\
\midrule
\multirow{6}{*}{\textbf{BLEU}$\uparrow$} & T=100  & \textbf{88.86}  & 87.53 ($\downarrow$ 1.33)   &87.36 ($\downarrow$ 1.50)                                     \\
                               & T=500                     & \textbf{88.68}           & 86.43 ($\downarrow$ 2.25)  & 87.26 ($\downarrow$ 1.42)                                    \\
                               & T=1000                          & \textbf{87.64}           & 85.52 ($\downarrow$ 2.12)  &86.72 ($\downarrow$ 0.92)                                     \\
                               & T=1500                         & \textbf{87.36}           & 85.32 ($\downarrow$ 2.04)    &86.56 ($\downarrow$ 0.80)                                    \\ 
                               & T=2000      &    \textbf{87.37}   &   85.31 ($\downarrow$ 2.06) &  86.55 ($\downarrow$ 0.82)  \\
                               & T=2500      &    \textbf{87.40}   &   85.33 ($\downarrow$ 2.07) &  86.63 ($\downarrow$ 0.77)  \\
\midrule
\multirow{6}{*}{\textbf{PPL}$\downarrow$}  & T=100                         & \textbf{4.74}             & 5.04 ($\uparrow$ 0.30)     &4.87 ($\uparrow$ 0.13)                                     \\
                               & T=500                         & \textbf{4.53}             & 5.04 ($\uparrow$ 0.51)     &4.80 ($\uparrow$ 0.27)                                     \\
                               & T=1000                         & \textbf{4.63}             & 5.07 ($\uparrow$ 0.44)     & 4.78 ($\uparrow$ 0.15)                                     \\
                               & T=1500                         & \textbf{4.69}             & 5.09 ($\uparrow$ 0.40)     &4.78 ($\uparrow$ 0.09)                                     \\ 
                               & T=2000      &   \textbf{4.68}     &    5.07 ($\uparrow$ 0.39)    &   4.74 ($\uparrow$ 0.06)  \\
                               & T=2500      &   \textbf{4.67}     &    5.06 ($\uparrow$ 0.39)    &   4.81 ($\uparrow$ 0.14)  \\
\bottomrule
\end{tabular}
\end{table}

\begin{table}[t]
\centering
\caption{BLEU scores and PPL on previous edits after the last step editing (i.e., 2500 instances). Each score denotes the result (the average of four ranks) on the multi-rank queries of each previous step. The best results are highlighted with \textbf{Bold}.}
\label{tab:history_eval}
\begin{tabular}{c|c|cc}
\toprule
\multicolumn{1}{c|}{\textbf{Metric}}     & \multicolumn{1}{c|}{\textbf{\#Edits}} & \textbf{EvoEdit} & \multicolumn{1}{c}{\textbf{w/o KPF}} \\
\midrule
\multirow{6}{*}{\textbf{BLEU}$\uparrow$} & T=100  & \textbf{37.47}   &34.22 ($\downarrow$ 3.25)                                     \\
                               & T=500                     & \textbf{40.56}           & 36.84 ($\downarrow$ 3.72)                                    \\
                               & T=1000                          & \textbf{41.86}          & 37.67 ($\downarrow$ 4.19)                                     \\
                               & T=1500                         & \textbf{37.77}              &34.74 ($\downarrow$ 3.03)                                    \\ 
                               & T=2000      &    \textbf{40.81}   &  37.02 ($\downarrow$ 3.79)  \\
                               & T=2450      &    \textbf{41.72}    &  34.35 ($\downarrow$ 7.37)  \\
\midrule
\multirow{6}{*}{\textbf{PPL}$\downarrow$}  & T=100                         & \textbf{83.65}          &174.51 ($\uparrow$ 90.86)             \\
                               & T=500                         & \textbf{54.79}            &117.59 ($\uparrow$ 62.80)                                     \\
                               & T=1000                         & \textbf{53.72}            & 118.84 ($\uparrow$ 65.12)                                     \\
                               & T=1500                         & \textbf{65.53}              &149.29 ($\uparrow$ 83.76)                                     \\ 
                               & T=2000      &   \textbf{47.14}         &   108.83 ($\uparrow$ 61.69)  \\
                               & T=2450      &   \textbf{35.31}        &  131.16 ($\uparrow$ 95.85)  \\
\bottomrule
\end{tabular}
\end{table}

\begin{figure*}[t]
    \centering
    \includegraphics[width=1.0\textwidth]{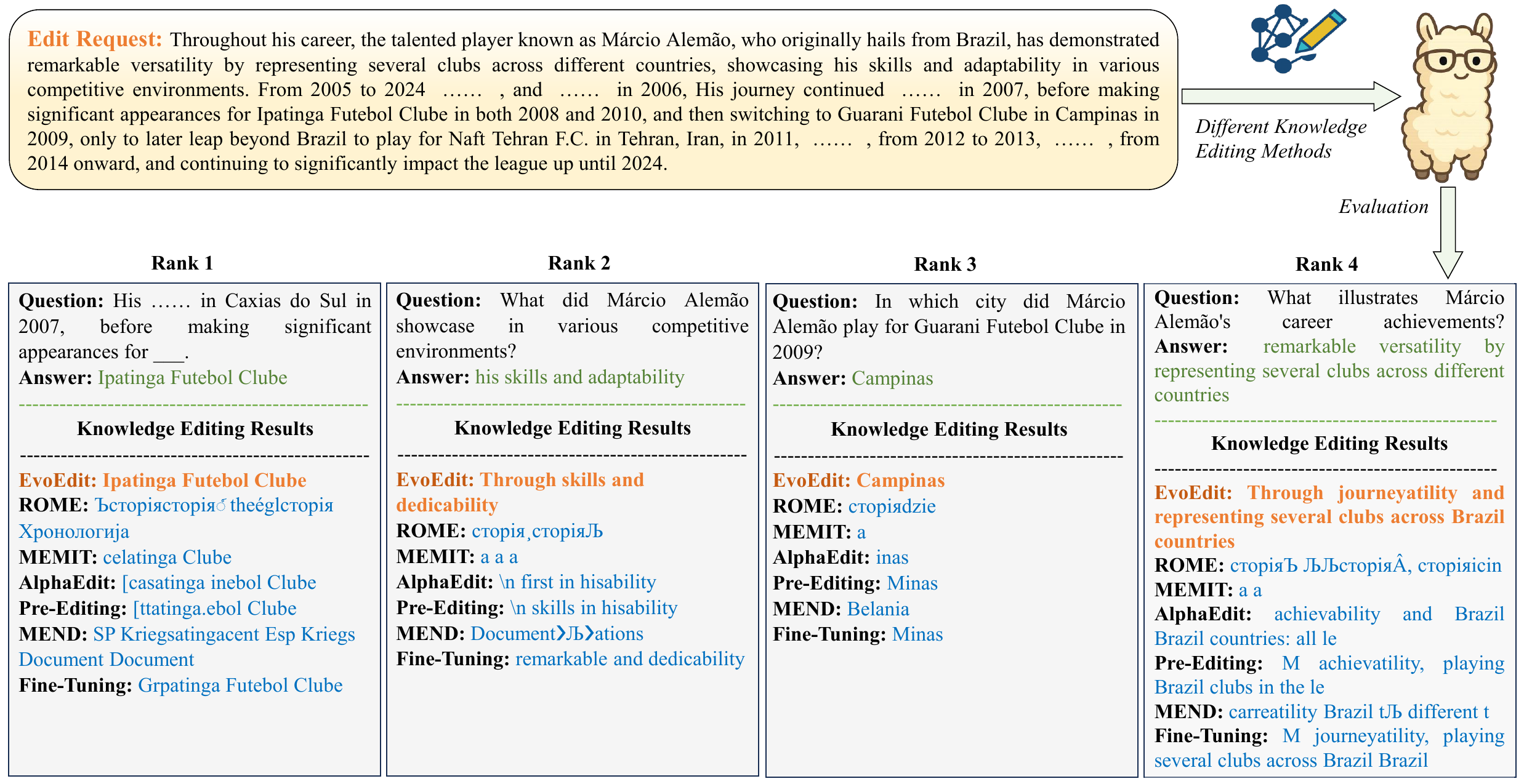}
    \caption{An editing example to qualitatively illustrate the effectiveness of the proposed method \textsc{EvoEdit} for the LF-Edit task.}
    \label{fig:case} 
\end{figure*}

\subsection{Ablation Study} \label{AS}
To demonstrate the effectiveness of latent perturbation augmentation (LPA) and knowledge-driven parameter fusion (KPF), we conduct an ablation study for answering the research question \textbf{RQ3}. We ablate the LPA and KPF separately to create two variants: 1) w/o LPA, which removes the latent perturbation augmentation module; 2) w/o KPF, which does not merge previous parameters (i.e., original parameters and the last step parameters) after each knowledge editing. The results of the ablation study are presented in Table \ref{tab:ablation_study_on_modules}. We further assess KPF’s ability to preserve previously edited knowledge in Table~\ref{tab:history_eval}, and compare alternative parameter-merging strategies in Figure~\ref{parameter-merging}. From the results, we can conclude that: 

\textbf{1) Effectiveness of Latent Perturbation Augmentation.} To evaluate the impact of the LPA module, we remove it from the proposed method \textsc{EvoEdit}. When editing 500 instances (i.e., T = 500), the BLEU score on MRLF-Bench decreases by 2.25 points. Similarly, the score drops by 2.12 and 2.04 points when editing 1000 and 1500 instances, respectively. Moreover, the removal of the LPA module leads to higher perplexity in the model’s outputs. These performance degradations demonstrate that the LPA module effectively enhances \textsc{EvoEdit}’s ability to perform lifelong free-text knowledge editing.

\textbf{2) Effectiveness of Knowledge-driven Parameter Fusion.} As shown in Table \ref{tab:ablation_study_on_modules}, removing the KPF module results in BLEU score reductions of 1.50 and 1.42 when editing 100 and 500 instances, respectively. To further assess the impact of parameter fusion, we measure the model’s PPL on previously learned knowledge. As illustrated in Table \ref{tab:history_eval}, eliminating the KPF module causes a substantial performance degradation for prior edits, indicating that KPF helps the model better retain and consolidate previously acquired knowledge. In addition, we also compare KPF with the direct parameter fusion (DPF) strategy, as shown in Figure \ref{parameter-merging}. The substantial performance improvement proves the effectiveness of the proposed KPF module for this new task.

% \begin{figure}[ht]
%     \centering
%     \includegraphics[width=\columnwidth]{perplexity_for_ablation.png}
%     \caption{Ablation study by removing the main components on the LLaMA-3. We calculate the PPL of the model to the previously learned knowledge under different evaluation settings.}
%     \label{fig:perplexity_for_ablation}
% \end{figure}

\subsection{Case Study} \label{case study}
We present an illustrative example in Figure \ref{fig:case} to demonstrate the effectiveness of the proposed method, \textsc{EvoEdit}, for the LF-Edit task, in comparison with other knowledge-editing baselines. From the results, we observe the following: 1) Accuracy in Knowledge Integration: Compared with the baselines, our method consistently produces correct answers across all rank levels. For simpler queries, \textsc{EvoEdit} generates predictions that exactly match the ground truth, while for more complex cases, it produces semantically equivalent responses. This demonstrates that our approach effectively internalizes newly introduced knowledge. 2) Fluency in Natural Language Generation: \textsc{EvoEdit} also yields smoother and more natural language outputs. In contrast, other knowledge-editing methods often fail to produce coherent or contextually appropriate text, sometimes even compromising standard text generation abilities. This issue arises because continual editing significantly disrupts the model’s fundamental language capabilities, and the accumulated negative effects can lead to model degradation. Hence, preserving the model’s core linguistic competence throughout the editing process is essential.

\section{Conclusion and Future Work}
In this paper, we introduce a new task termed lifelong free-text knowledge editing, which enables large language models (LLMs) to continuously acquire new knowledge from free-text inputs while retaining previously learned information without catastrophic forgetting. To support research in this direction, we develop MRLF-Bench, a multi-rank benchmark designed for evaluating and advancing this task. Furthermore, we propose \textsc{EvoEdit}, a method tailored for lifelong free-text knowledge editing. \textsc{EvoEdit} employs feature perturbation for data augmentation and selectively fuses parameters to balance the integration of new knowledge with the preservation of prior information. Experimental results demonstrate that our approach consistently surpasses existing baselines in both new knowledge acquisition and old knowledge retention. In the future, we could extend this work from two perspectives: 1) for the data construction, the The length of free text can be longer and contain more types of knowledge, not just limited to factual knowledge. 2) for the editing method, we need to develop knowledge editing algorithms that are suitable for larger and more diverse language models.

\bibliographystyle{IEEEtran}
\bibliography{references}

@misc{zhang2024comprehensivestudyknowledgeediting,
      title={A Comprehensive Study of Knowledge Editing for Large Language Models}, 
      author={Ningyu Zhang and Yunzhi Yao and Bozhong Tian and Peng Wang and Shumin Deng and Mengru Wang and Zekun Xi and Shengyu Mao and Jintian Zhang and Yuansheng Ni and Siyuan Cheng and Ziwen Xu and Xin Xu and Jia-Chen Gu and Yong Jiang and Pengjun Xie and Fei Huang and Lei Liang and Zhiqiang Zhang and Xiaowei Zhu and Jun Zhou and Huajun Chen},
      year={2024},
      eprint={2401.01286},
      archivePrefix={arXiv},
}

@inproceedings{cheng2024editmultimodallargelanguage,
  title={Can We Edit Multimodal Large Language Models?},
  author={Cheng, Siyuan and Tian, Bozhong and Liu, Qingbin and Chen, Xi and Wang, Yongheng and Chen, Huajun and Zhang, Ningyu},
  booktitle={Proceedings of the 2023 Conference on Empirical Methods in Natural Language Processing},
  pages={13877--13888},
  year={2023}
}

@inproceedings{survey_for_knowledge_editing,
  title={Editing Large Language Models: Problems, Methods, and Opportunities},
  author={Yao, Yunzhi and Wang, Peng and Tian, Bozhong and Cheng, Siyuan and Li, Zhoubo and Deng, Shumin and Chen, Huajun and Zhang, Ningyu},
  booktitle={Proceedings of the 2023 Conference on Empirical Methods in Natural Language Processing},
  pages={10222--10240},
  year={2023}
}

@misc{survet_of_finetuning,
      title={Parameter-Efficient Fine-Tuning for Foundation Models}, 
      author={Dan Zhang and Tao Feng and Lilong Xue and Yuandong Wang and Yuxiao Dong and Jie Tang},
      year={2025},
      eprint={2501.13787},
      archivePrefix={arXiv},
}

@misc{survey_of_RAG,
      title={Synergizing RAG and Reasoning: A Systematic Review}, 
      author={Yunfan Gao and Yun Xiong and Yijie Zhong and Yuxi Bi and Ming Xue and Haofen Wang},
      year={2025},
      eprint={2504.15909},
      archivePrefix={arXiv},
}

@article{rome,
  title={Locating and editing factual associations in gpt},
  author={Meng, Kevin and Bau, David and Andonian, Alex and Belinkov, Yonatan},
  journal={Advances in neural information processing systems},
  volume={35},
  pages={17359--17372},
  year={2022}
}

@inproceedings{memit,
  title={Mass-Editing Memory in a Transformer},
  author={Meng, Kevin and Sharma, Arnab Sen and Andonian, Alex J and Belinkov, Yonatan and Bau, David},
  booktitle={ICLR},
  year={2023}
}

@inproceedings{alphaedit,
  title={AlphaEdit: Null-Space Constrained Knowledge Editing for Language Models},
  author={Fang, Junfeng and Jiang, Houcheng and Wang, Kun and Ma, Yunshan and Shi, Jie and Wang, Xiang and He, Xiangnan and Chua, Tat-Seng},
  booktitle={The Thirteenth International Conference on Learning Representations}
}

@inproceedings{mend,
  title={Fast Model Editing at Scale},
  author={Mitchell, Eric and Lin, Charles and Bosselut, Antoine and Finn, Chelsea and Manning, Christopher D},
  booktitle={International Conference on Learning Representations}
}

@inproceedings{serac,
  title={Memory-based model editing at scale},
  author={Mitchell, Eric and Lin, Charles and Bosselut, Antoine and Manning, Christopher D and Finn, Chelsea},
  booktitle={International Conference on Machine Learning},
  pages={15817--15831},
  year={2022},
  organization={PMLR}
}

@article{wise,
  title={Wise: Rethinking the knowledge memory for lifelong model editing of large language models},
  author={Wang, Peng and Li, Zexi and Zhang, Ningyu and Xu, Ziwen and Yao, Yunzhi and Jiang, Yong and Xie, Pengjun and Huang, Fei and Chen, Huajun},
  journal={Advances in Neural Information Processing Systems},
  volume={37},
  pages={53764--53797},
  year={2024}
}

@inproceedings{r-ROME,
  title={Rebuilding ROME: Resolving Model Collapse during Sequential Model Editing},
  author={Gupta, Akshat and Baskaran, Sidharth and Anumanchipalli, Gopala},
  booktitle={Proceedings of the 2024 Conference on Empirical Methods in Natural Language Processing},
  pages={21738--21744},
  year={2024}
}

@inproceedings{anyedit,
  title={AnyEdit: Edit Any Knowledge Encoded in Language Models},
  author={Jiang, Houcheng and Fang, Junfeng and Zhang, Ningyu and Wan, Mingyang and Ma, Guojun and Wang, Xiang and He, Xiangnan and Chua, Tat-Seng},
  booktitle={Forty-second International Conference on Machine Learning}
}

@inproceedings{dem,
  title={Commonsense Knowledge Editing Based on Free-Text in LLMs},
  author={Huang, Xiusheng and Wang, Yequan and Zhao, Jun and Liu, Kang},
  booktitle={Proceedings of the 2024 Conference on Empirical Methods in Natural Language Processing},
  pages={14870--14880},
  year={2024}
}

@inproceedings{wilke,
  title={WilKE: Wise-Layer Knowledge Editor for Lifelong Knowledge Editing},
  author={Hu, Chenhui and Cao, Pengfei and Chen, Yubo and Liu, Kang and Zhao, Jun},
  booktitle={Findings of the Association for Computational Linguistics ACL 2024},
  pages={3476--3503},
  year={2024}
}

@misc{memla,
      title={MEMLA: Enhancing Multilingual Knowledge Editing with Neuron-Masked Low-Rank Adaptation}, 
      author={Jiakuan Xie and Pengfei Cao and Yuheng Chen and Yubo Chen and Kang Liu and Jun Zhao},
      year={2024},
      eprint={2406.11566},
      archivePrefix={arXiv},
}

@misc{fine,
      title={Precise Localization of Memories: A Fine-grained Neuron-level Knowledge Editing Technique for LLMs}, 
      author={Haowen Pan and Xiaozhi Wang and Yixin Cao and Zenglin Shi and Xun Yang and Juanzi Li and Meng Wang},
      year={2025},
      eprint={2503.01090},
      archivePrefix={arXiv},
}

@inproceedings{ke,
  title={Editing Factual Knowledge in Language Models},
  author={De Cao, Nicola and Aziz, Wilker and Titov, Ivan},
  booktitle={Proceedings of the 2021 Conference on Empirical Methods in Natural Language Processing},
  pages={6491--6506},
  year={2021}
}

@inproceedings{malman,
  title={Massive Editing for Large Language Models via Meta Learning},
  author={Tan, Chenmien and Zhang, Ge and Fu, Jie},
  booktitle={The Twelfth International Conference on Learning Representations}
}

@inproceedings{ike,
  title={Can We Edit Factual Knowledge by In-Context Learning?},
  author={Zheng, Ce and Li, Lei and Dong, Qingxiu and Fan, Yuxuan and Wu, Zhiyong and Xu, Jingjing and Chang, Baobao},
  booktitle={Proceedings of the 2023 Conference on Empirical Methods in Natural Language Processing},
  pages={4862--4876},
  year={2023}
}

@article{grace,
  title={Aging with grace: Lifelong model editing with discrete key-value adaptors},
  author={Hartvigsen, Tom and Sankaranarayanan, Swami and Palangi, Hamid and Kim, Yoon and Ghassemi, Marzyeh},
  journal={Advances in Neural Information Processing Systems},
  volume={36},
  pages={47934--47959},
  year={2023}
}

@inproceedings{mello,
  title={MQuAKE: Assessing Knowledge Editing in Language Models via Multi-Hop Questions},
  author={Zhong, Zexuan and Wu, Zhengxuan and Manning, Christopher D and Potts, Christopher and Chen, Danqi},
  booktitle={Proceedings of the 2023 Conference on Empirical Methods in Natural Language Processing},
  pages={15686--15702},
  year={2023}
}

@Inbook{Piaget,
author="Bovet, Magali",
editor="Inhelder, B{\"a}rbel
and Chipman, Harold H.
and Zwingmann, Charles",
title="Piaget's Theory of Cognitive Development and Individual Differences",
bookTitle="Piaget and His School: A Reader in Developmental Psychology",
year="1976",
publisher="Springer Berlin Heidelberg",
address="Berlin, Heidelberg",
pages="269--279",
isbn="978-3-642-46323-5",
doi="10.1007/978-3-642-46323-5_20",
}

@inproceedings{zsre,
  title={Zero-Shot Relation Extraction via Reading Comprehension},
  author={Levy, Omer and Seo, Minjoon and Choi, Eunsol and Zettlemoyer, Luke},
  booktitle={Proceedings of the 21st Conference on Computational Natural Language Learning (CoNLL 2017)},
  pages={333--342},
  year={2017}
}

@inproceedings{mulfe,
    title = "{MULFE}: A Multi-Level Benchmark for Free Text Model Editing",
    author = "Wang, Chenhao  and
      Cao, Pengfei  and
      Jin, Zhuoran  and
      Chen, Yubo  and
      Zeng, Daojian  and
      Liu, Kang  and
      Zhao, Jun",
    editor = "Ku, Lun-Wei  and
      Martins, Andre  and
      Srikumar, Vivek",
    booktitle = "Proceedings of the 62nd Annual Meeting of the Association for Computational Linguistics (Volume 1: Long Papers)",
    month = aug,
    year = "2024",
    address = "Bangkok, Thailand",
    publisher = "Association for Computational Linguistics",
    doi = "10.18653/v1/2024.acl-long.732",
    pages = "13570--13587"
}

@inproceedings{stanford-openie,
    title = "Leveraging Linguistic Structure For Open Domain Information Extraction",
    author = "Angeli, Gabor  and
      Johnson Premkumar, Melvin Jose  and
      Manning, Christopher D.",
    editor = "Zong, Chengqing  and
      Strube, Michael",
    booktitle = "Proceedings of the 53rd Annual Meeting of the Association for Computational Linguistics and the 7th International Joint Conference on Natural Language Processing (Volume 1: Long Papers)",
    month = jul,
    year = "2015",
    address = "Beijing, China",
    publisher = "Association for Computational Linguistics",
    doi = "10.3115/v1/P15-1034",
    pages = "344--354"
}

@inproceedings{bleu,
    title = "{B}leu: a Method for Automatic Evaluation of Machine Translation",
    author = "Papineni, Kishore  and
      Roukos, Salim  and
      Ward, Todd  and
      Zhu, Wei-Jing",
    editor = "Isabelle, Pierre  and
      Charniak, Eugene  and
      Lin, Dekang",
    booktitle = "Proceedings of the 40th Annual Meeting of the Association for Computational Linguistics",
    month = jul,
    year = "2002",
    address = "Philadelphia, Pennsylvania, USA",
    publisher = "Association for Computational Linguistics",
    doi = "10.3115/1073083.1073135",
    pages = "311--318"
}

@misc{survey_for_extraction,
      title={A Survey on Neural Open Information Extraction: Current Status and Future Directions}, 
      author={Shaowen Zhou and Bowen Yu and Aixin Sun and Cheng Long and Jingyang Li and Haiyang Yu and Jian Sun and Yongbin Li},
      year={2022},
      eprint={2205.11725},
      archivePrefix={arXiv},
}

@misc{RegMean,
      title={Dataless Knowledge Fusion by Merging Weights of Language Models}, 
      author={Xisen Jin and Xiang Ren and Daniel Preotiuc-Pietro and Pengxiang Cheng},
      year={2025},
      eprint={2212.09849},
      archivePrefix={arXiv},
}

@inproceedings{kecontinual,
  title={Continual Pre-training of Language Models},
  author={Ke, Zixuan and Shao, Yijia and Lin, Haowei and Konishi, Tatsuya and Kim, Gyuhak and Liu, Bing},
  booktitle={The Eleventh International Conference on Learning Representations}
}

@inproceedings{hu2024wilke,
  title={WilKE: Wise-Layer Knowledge Editor for Lifelong Knowledge Editing},
  author={Hu, Chenhui and Cao, Pengfei and Chen, Yubo and Liu, Kang and Zhao, Jun},
  booktitle={Findings of the Association for Computational Linguistics ACL 2024},
  pages={3476--3503},
  year={2024}
}

@inproceedings{wang2024mulfe,
  title={MULFE: A multi-level benchmark for free text model editing},
  author={Wang, Chenhao and Cao, Pengfei and Jin, Zhuoran and Chen, Yubo and Zeng, Daojian and Liu, Kang and Zhao, Jun},
  booktitle={Proceedings of the 62nd Annual Meeting of the Association for Computational Linguistics (Volume 1: Long Papers)},
  pages={13570--13587},
  year={2024}
}

@inproceedings{jin2024cutting,
  title={Cutting Off the Head Ends the Conflict: A Mechanism for Interpreting and Mitigating Knowledge Conflicts in Language Models},
  author={Jin, Zhuoran and Cao, Pengfei and Yuan, Hongbang and Chen, Yubo and Xu, Jiexin and Li, Huaijun and Jiang, Xiaojian and Liu, Kang and Zhao, Jun},
  booktitle={Findings of the Association for Computational Linguistics ACL 2024},
  pages={1193--1215},
  year={2024}
}

@inproceedings{hu2025knowledge,
  title={Knowledge in superposition: Unveiling the failures of lifelong knowledge editing for large language models},
  author={Hu, Chenhui and Cao, Pengfei and Chen, Yubo and Liu, Kang and Zhao, Jun},
  booktitle={Proceedings of the AAAI Conference on Artificial Intelligence},
  volume={39},
  number={22},
  pages={24086--24094},
  year={2025}
}

@inproceedings{ilharcoediting,
  title={Editing models with task arithmetic},
  author={Ilharco, Gabriel and Ribeiro, Marco Tulio and Wortsman, Mitchell and Schmidt, Ludwig and Hajishirzi, Hannaneh and Farhadi, Ali},
  booktitle={The Eleventh International Conference on Learning Representations}
}

@article{akiba2025evolutionary,
  title={Evolutionary optimization of model merging recipes},
  author={Akiba, Takuya and Shing, Makoto and Tang, Yujin and Sun, Qi and Ha, David},
  journal={Nature Machine Intelligence},
  volume={7},
  number={2},
  pages={195--204},
  year={2025},
  publisher={Nature Publishing Group UK London}
}

@article{liu2024checkpoint,
  title={Checkpoint merging via bayesian optimization in llm pretraining},
  author={Liu, Deyuan and Wang, Zecheng and Wang, Bingning and Chen, Weipeng and Li, Chunshan and Tu, Zhiying and Chu, Dianhui and Li, Bo and Sui, Dianbo},
  journal={arXiv preprint arXiv:2403.19390},
  year={2024}
}

@inproceedings{yangadamerging,
  title={AdaMerging: Adaptive Model Merging for Multi-Task Learning},
  author={Yang, Enneng and Wang, Zhenyi and Shen, Li and Liu, Shiwei and Guo, Guibing and Wang, Xingwei and Tao, Dacheng},
  booktitle={The Twelfth International Conference on Learning Representations}
}

@article{matena2022merging,
  title={Merging models with fisher-weighted averaging},
  author={Matena, Michael S and Raffel, Colin A},
  journal={Advances in Neural Information Processing Systems},
  volume={35},
  pages={17703--17716},
  year={2022}
}

@article{yadav2023resolving,
  title={Resolving interference when merging models},
  author={Yadav, Prateek and Tam, Derek and Choshen, Leshem and Raffel, Colin and Bansal, Mohit},
  journal={arXiv preprint arXiv:2306.01708},
  volume={1},
  year={2023}
}

@inproceedings{zhou2024metagpt,
  title={MetaGPT: Merging Large Language Models Using Model Exclusive Task Arithmetic},
  author={Zhou, Yuyan and Song, Liang and Wang, Bingning and Chen, Weipeng},
  booktitle={Proceedings of the 2024 Conference on Empirical Methods in Natural Language Processing},
  pages={1711--1724},
  year={2024}
}

@article{tang2023concrete,
  title={Concrete subspace learning based interference elimination for multi-task model fusion},
  author={Tang, Anke and Shen, Li and Luo, Yong and Ding, Liang and Hu, Han and Du, Bo and Tao, Dacheng},
  journal={arXiv preprint arXiv:2312.06173},
  year={2023}
}

@inproceedings{yu2024language,
  title={Language Models are Super Mario: Absorbing Abilities from Homologous Models as a Free Lunch},
  author={Yu, Le and Yu, Bowen and Yu, Haiyang and Huang, Fei and Li, Yongbin},
  booktitle={International Conference on Machine Learning},
  pages={57755--57775},
  year={2024},
  organization={PMLR}
}

@inproceedings{zhu2024model,
  title={Model Tailor: Mitigating Catastrophic Forgetting in Multi-modal Large Language Models},
  author={Zhu, Didi and Sun, Zhongyisun and Li, Zexi and Shen, Tao and Yan, Ke and Ding, Shouhong and Wu, Chao and Kuang, Kun},
  booktitle={International Conference on Machine Learning},
  pages={62581--62598},
  year={2024},
  organization={PMLR}
}

@inproceedings{wang2024localizing,
  title={Localizing Task Information for Improved Model Merging and Compression},
  author={Wang, Ke and Dimitriadis, Nikolaos and Ortiz-Jimenez, Guillermo and Fleuret, Fran{\c{c}}ois and Frossard, Pascal},
  booktitle={International Conference on Machine Learning},
  pages={50268--50287},
  year={2024},
  organization={PMLR}
}

@article{huang2024emr,
  title={Emr-merging: Tuning-free high-performance model merging},
  author={Huang, Chenyu and Ye, Peng and Chen, Tao and He, Tong and Yue, Xiangyu and Ouyang, Wanli},
  journal={Advances in Neural Information Processing Systems},
  volume={37},
  pages={122741--122769},
  year={2024}
}

@article{cao2024one,
  title={One Mind, Many Tongues: A Deep Dive into Language-Agnostic Knowledge Neurons in Large Language Models},
  author={Cao, Pengfei and Chen, Yuheng and Jin, Zhuoran and Chen, Yubo and Liu, Kang and Zhao, Jun},
  journal={arXiv preprint arXiv:2411.17401},
  year={2024}
}

\end{document}